# Statistical learning for accurate and interpretable battery lifetime prediction

Peter M. Attia[1*], Kristen A. Severson[2], Jeremy D. Witmer[3]

1. Department of Materials Science and Engineering, Stanford University, Stanford, CA, USA
2. MIT-IBM Watson AI Lab, IBM Research, Cambridge, MA, USA
3. Ginzton Laboratory, Stanford University, Stanford, CA, USA

*Correspondence: peter.m.attia@gmail.com

## Abstract

Data-driven methods for battery lifetime prediction are attracting increasing attention for applications in which the degradation mechanisms are poorly understood and suitable training sets are available. However, while advanced machine learning and deep learning methods promise high performance with minimal data preprocessing, simpler linear models with engineered features often achieve comparable performance, especially for small training sets, while also providing physical and statistical interpretability. In this work, we use a previously published dataset to develop simple, accurate, and interpretable data-driven models for battery lifetime prediction. We first present the "capacity matrix" concept as a compact representation of battery electrochemical cycling data, along with a series of feature representations. We then create a number of univariate and multivariate models, many of which achieve comparable performance to the highest-performing models previously published for this dataset. These models also provide insights into the degradation of these cells. Our approaches can be used both to quickly train models for a new dataset and to benchmark the performance of more advanced machine learning methods.

## Graphical abstract

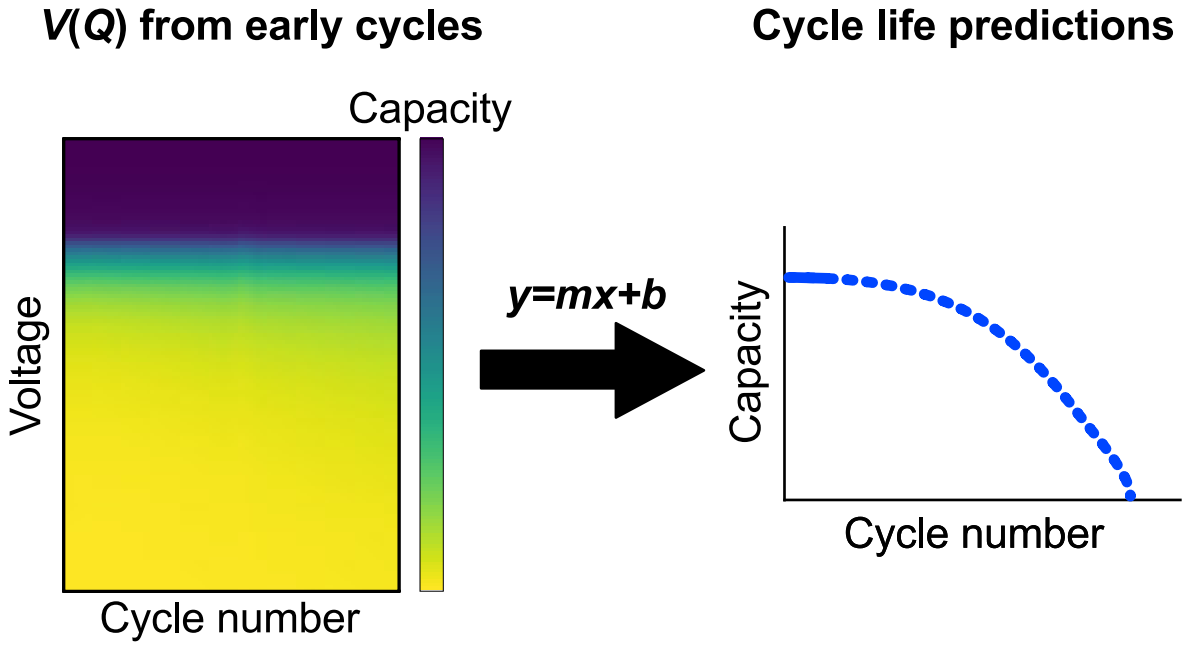

Battery lifetime prediction has many applications throughout the battery product cycle. Some examples include screening of new electrodes, electrolytes, and cell designs during research and development; optimizing cell designs, cycling protocols for formation (the final step in cell manufacturing), and fast charging protocols; and predicting cycle life, remaining capacity, and the likelihood of a safety-threatening event in the context of use and reuse in the field. Critically, accurate battery lifetime estimation is needed to set warranty cost estimates for electric vehicle and grid storage applications, as reducing the uncertainty in the warranty cost will reduce the cost of battery deployments. Given the many applications of lifetime prediction within the battery product cycle, this research direction is of crucial importance to the battery community. While improving our first-principles understanding and modeling of battery degradation is an imperative research direction[1,2], data-driven approaches to battery lifetime prediction[3–10] are increasingly exciting for applications in which the degradation mechanisms are challenging to model and a suitable training dataset is available.

Desirable attributes of data-driven models for battery lifetime prediction include high accuracy, low number of cycles required for prediction, small training set sizes, and high interpretability. The first three of these criteria are attributes related to model performance, and the final criterion engenders trust in the model. The use of sophisticated machine learning methods is undoubtedly an exciting research direction, as these methods can produce high performing models with minimal domain expertise (which is required to design predictive features). In particular, advanced machine learning and deep learning methodologies often excel at capturing information from high-dimensional datasets[11], which in many ways makes them an ideal fit for high-dimensional battery cycling datasets, i.e., voltage vs. capacity as a function of cycle number. However, deep learning models are often brittle, i.e., small changes in input data can lead to dramatic differences in their predictions[12,13], and generally require large training set sizes.[11] Furthermore, these methods often produce models that are not easily interpretable, as the relationship between the input and output data is convoluted. Interpretability helps confirm the model is behaving reasonably and, in some situations, can even elucidate the underlying physics of the task at hand. Thus, interpretability is a useful property for machine learning models applied to scientific domains, particularly for experimentally generated datasets (which unavoidably have "real-world" issues). While developing frameworks to explain advanced machine learning and deep learning methods is an active area of research[14,15], explaining black box models has inherent limitations compared to using intrinsically interpretable approaches.[16] For instance, linear models using "engineered" (i.e., curated) input features that are downselected via regularization[17] are often interpretable while still achieving high predictive performance, especially with small- or medium-sized datasets. Throughout this work, we refer to classical methods that produce these types of models as "statistical learning" methods.[17] Models generated via statistical learning methods can be trained quickly and can be used to benchmark the performance of models generated via more complex methods. Additionally, the interpretability of these models can aid in generalizing across training sets and in developing physics-informed data-driven models[18,19]. While different data-driven approaches may be useful in different contexts, statistical learning for battery lifetime prediction remains underexplored in the literature.

Severson et al.[3] developed statistical learning models to predict the lifetime of lithium iron phosphate (LFP)/graphite cylindrical cells undergoing ~10-minute fast charging using a training dataset of 41 cells cycled to failure. Degradation during fast charging in commercially-relevant form factors is poorly understood[20,21] and thus challenging to model using first principles, making data-driven approaches a suitable alternative given an available training set. These machine learning models, which were subsequently used for rapid evaluation of fast-charging protocols[22], achieved ~9% test error using only the first 100 cycles for prediction (12% of the average cycle life). The features, i.e., transformations of the raw data, were sourced from measurements of voltage vs. capacity, capacity vs. cycle number, internal resistance, and can temperature. The most predictive features were derived from the voltage vs. discharge capacity curves, which is among the richest yet most complex data sources; specifically, these features summarized the difference between the voltage vs. discharge capacity curves of the 100$^{th}$ and 10$^{th}$ cycles, denoted $\Delta Q_{100-10}(V)$. In fact, a simple linear model with a single feature derived from $\Delta Q_{100-10}(V)$ performed nearly as well as the most complex model containing information from all data sources. This result highlighted the value of using features from the voltage vs. capacity data, as opposed to just the capacity vs. cycle number data. More importantly, the high performance of these simple models demonstrated the power of domain-inspired feature engineering coupled with statistical learning methods.

Since publication of the Severson et al.[3] dataset, others have applied advanced machine learning and deep learning methods, including relevance vector machines[5,8], gradient boosted regression trees[23], Gaussian process regression[8], recurrent neural networks (including long short-term memory networks)[8,24], and convolutional neural networks[8,25–28]. Many of these works have explored creative approaches, including data augmentation[25] and the use of differential capacity analysis[28]. However, few of these approaches emphasize interpretability. Our belief is that the development of interpretable statistical learning approaches for battery lifetime prediction should be pursued in tandem with state-of-the-art machine learning and deep learning methods that maximize performance.

In this work, we explore statistical learning approaches for developing accurate and interpretable battery lifetime prediction models. We use the same datasets and objectives in Severson et al.[3] for consistency and for comparison. We first present the "capacity matrix" concept for compactly representing the changes in cell capacity with respect to voltage and cycle number. We then present a number of dimensionality reduction and model building approaches and apply them to this dataset. While models as simple as a univariate linear model using a single element of $\Delta Q_{100-10}(V)$ perform similarly to the best models from Severson et al.[3], new approaches using the elements of $\Delta Q_{100-10}(V)$ further reduce error and provide interesting insights into the relationship between elements in $\Delta Q_{100-10}(V)$ and cycle life. Overall, the high accuracy of these simple, interpretable models highlights the effectiveness of statistical learning. Additionally, features sourced from other transformations of the capacity matrix are generally outperformed by features from $\Delta Q_{100-10}(V)$; throughout this work, we discuss both successful and unsuccessful approaches to reduce prediction error. The methods in this work can apply broadly to other battery datasets and can serve as a benchmarking suite for new datasets.

## Summary of dataset and previous work

Here, we summarize the dataset and statistical learning approach in Severson et al.[3], but we refer the reader to the original publication for more information. Table I also summarizes essential details of this study. The dataset consists of 124 LFP/graphite cells cycled in three "batches", i.e., a group of cells cycled simultaneously in the convection oven. These cells were split into a training set (41 cells), a primary test set (43 cells), and a secondary test set (40 cells; generated after model development). Cells were cycled with one of 72 different fast charging protocols (~9–13 minutes from 0–80% state-of-charge), but all cells were identically discharged at 4C (here, C rate refers to the rate required to (dis)charge a cell in 1 hour). In addition to standard electrochemical data like voltage vs. capacity and capacity vs. cycle number, internal resistance was recorded every cycle number, and a thermocouple mounted to the cell can continuously recorded the can surface temperature of each cell.

**Table I.** Summary of the Severson et al.[3] dataset. See Severson et al.[3] for additional details.

| **Attribute** | **Description** |
|---|---|
| Cell chemistry (cathode/anode) | LFP/graphite |
| Cell manufacturer and type | A123 APR18650M1A |
| Nominal capacity | 1.1 Ah |
| Nominal voltage | 3.3 V |
| Voltage window | 2.0 V–3.6 V |
| Recommended fast charge current | 3.6C (4 A) |
| Number of cells used in study | 124 |
| Charge rate | Varied between 0–80% SOC; 1C CC-CV from 80% to 100% SOC |
| Discharge rate | 4C |
| Environmental temperature | 30°C |
| Temperature control method | Convection |

The modeling objective was to predict the log-transformed cycle life using data from only the first 100 cycles; this cycle number cutoff is notable because the capacity at this cycle number exceeds the initial capacity for most cells in the dataset (Severson et al.[3], Figure 1c; attributed to the diffusion of lithium from the edges to the center of the graphite electrode once cycling begins[29,30]). Cycle life was defined as the number of cycles before the capacity falls below 80% of the nominal capacity (80% of 1.1 Ah, or 0.88 Ah). The log transformation was applied to cycle life to reduce the positive skew of the cycle life distribution (Severson et al.[3], Supplementary Figure 28). Twenty features were developed from the various available data sources. The discharge curves were more suitable than the charge curves for features generated from voltage vs. capacity data because the discharge conditions were consistent across all cells; by using the discharge curves, the model is fitting to the intrinsic degradation of the cells and avoiding convolutions from the voltage response to the specific charging protocol. Regression models were developed using the elastic net, a regularized linear regression method that simultaneously selects a subset of features and determines the feature coefficients[31]. In general, the most-performant features were sourced from $\Delta Q_{100-10}(V)$; in fact, a model using only one such feature, $\log_{10}(\text{variance}(\Delta Q_{100-10}(V)))$, performed comparably to the "full" model using all available features (primary/secondary test errors of 138/196 cycles for the "variance" model vs. 100/214 cycles for the "full" model). The variance feature was hypothesized to correlate with the extent of nonuniformity in the dissipation of cell energy with respect to voltage; in other words, this energy dissipation becomes increasingly less uniform as the variance of $\Delta Q_{100-10}(V)$ increases. This initial nonuniformity may predict cycle life if the cell is degrading heterogeneously. We note that other features from $\Delta Q_{100-10}(V)$, including the mean and minimum, also performed well (although only the minimum was consistently selected during feature selection). Ultimately, the success of features from $\Delta Q_{100-10}(V)$ over features from the capacity fade curve was rationalized by the specific degradation mode (active material loss of the delithiated negative electrode), which initially affects the voltage response, but not the capacity, of the negative electrode.[32,33] This voltage response is also influenced by kinetic effects such as heterogeneous aging.[34]

## Matrix representation of voltage-capacity data

In Severson et al.[3], a model with a single feature extracted from the voltage vs. discharge capacity curves, $\log_{10}(\mathrm{var}(\Delta Q_{100-10}(V)))$, provided comparable accuracy to multivariate models with features from additional data sources such as internal resistance and can temperature. Clearly, the voltage vs. capacity curves contain important information for data-driven lifetime modeling in this application. While this model has not been extensively validated for other battery lifetime prediction tasks, Sulzer et al.[7] demonstrated that $\mathrm{var}(\Delta Q_{190-10}(V))$ from C/20 discharges is correlated with cycle life for NMC/graphite cells, which suggests that the variance feature generalizes to other datasets with more complex cell chemistries than LFP/graphite (note that this work uses the untransformed variance as opposed to the $\log_{10}$-transformed variance). In this work, we ignore the less predictive datastreams like internal resistance and temperature and focus on developing new representations of these voltage-capacity data. Another advantage of focusing on voltage-capacity data is that this data "comes for free" in a typical battery cycling experiment. In contrast, collecting data for other datastreams requires additional hardware and/or labor costs (e.g., can temperature measurements require manually attaching thermocouples to cell cans via thermal epoxy, using cell fixtures with embedded thermocouples, or a similar solution).

In principle, the relationship between voltage and capacity is continuous; in practice, this relationship is sampled at a finite rate during data acquisition. Severson et al.[3] fit each discharge voltage-capacity curve with a smoothing spline and then sampled each spline at 1000 evenly spaced points in voltage. Thus, for an operating window spanning 1.6 V (3.6–2.0 V), the distance between each capacity sample is 1.6 mV. Because the voltage window is constant with cycle number and capacity is changing with cycling, we fit capacity as a function of voltage (instead of voltage as a function of capacity) to ensure a consistent basis for comparison across cycle number. In fact, voltage as a function of capacity has been proposed and applied in previous work[5]. Note that linear interpolation is an alternative approach that also avoids the small additional error from the spline fit, but here we use the spline fits for consistency with Severson et al.[3] Lin et al.[35] suggest using a smoothing spline with a cross-validated smoothing parameter for sampling voltage-capacity curves, which is advantageous over manual tuning of filtering parameters. Alternatively, the data acquisition rate could be specified to match the desired basis vector.

Using this data processing approach, each discharge curve has 1000 capacity points at 1000 pre-defined voltage points. Thus, if we restrict the available training data to include only the first 100 cycles, each cell has 100,000 voltage-capacity data points. Of course, many of these data are highly correlated, both within a cycle and at the same voltage position across multiple cycles. As such, a primary goal of this work is to explore feature representations that reduce the dimensionality of these data without losing any information content.

We start by introducing a visualization of these 100,000 features, subsequently termed a "capacity matrix", that we found useful for model development. A similar approach has been developed by at least two previous works[36,37]. Figure 1 presents graphical representations of a capacity matrix. Figure 1a presents voltage vs. discharge capacity for the first 100 cycles for an example cell in the Severson et al.[3] dataset. Given the high discharge rates (4C), the voltage response is smoothed due to heterogeneity[38], and diagnostic differential capacity and/or differential voltage analysis would not reveal individual peaks that could be assigned to specific electrodes and failure modes. Overall, the curves shift only subtly as a function of cycle number. In Figure 1b, we present a capacity matrix representation of the data presented in Figure 1a, i.e., discharge capacity at 4C as a function of voltage and cycle number. A key advantage of this matrix representation is that this high-dimensional feature set is stored in a compact and machine-learning-ready format. However, the trends across cycle number are still challenging to perceive. Figure 1c presents the "baseline-subtracted capacity matrix", denoted $Q_n - Q_2$, which is simply the matrix presented in Figure 1b subtracted by the second column, i.e., the cycle 2 discharge capacity. We use cycle 2 here as the data from cycle 1 was unavailable due to a data acquisition error. With this representation, the subtle differences between cycles are much clearer. In Figure 1d, we present $Q_n - Q_2$ at two selected voltages, i.e., two rows of the baseline-subtracted capacity matrix. Curiously, the trends in these curves are quite linear, with the exception of the first 10 cycles (which we attribute to the diffusion of lithium from the edges to the center of the graphite electrode, commonly observed after an extended rest[29,30]) and cycles 50–60 (which we attribute to a temperature excursion in the environmental chamber during cycling of this batch). Other cycle numbers could also be selected for the baseline cycle; for instance, we use cycle 10 as a baseline cycle throughout this work as this cycle number avoids the initial rise in capacity seen in Figure 1d.

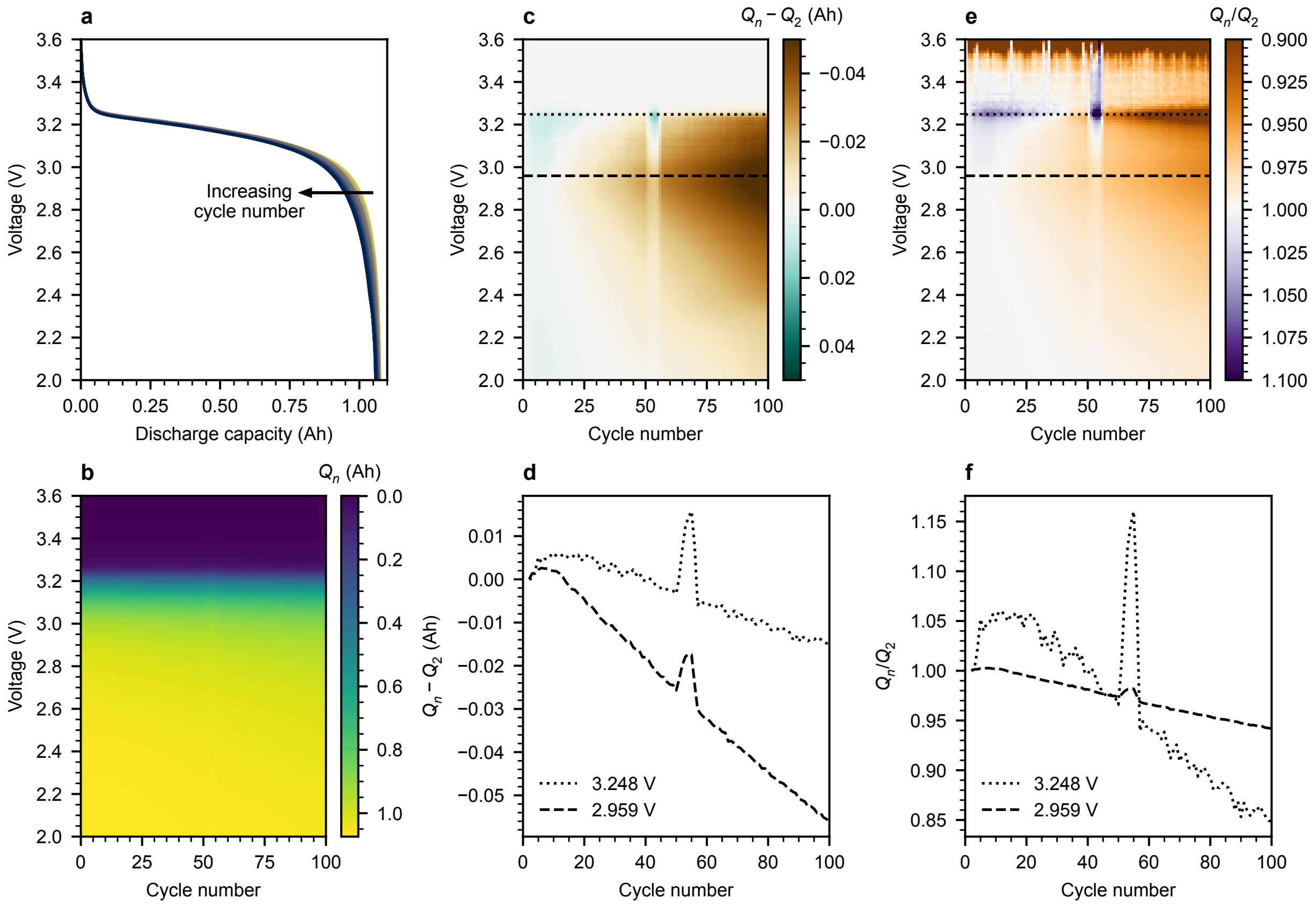

**Figure 1.** Graphical representations of a "capacity matrix", i.e., discharge capacity at 4C as a function of voltage and cycle number for an example cell in the Severson et al.[3] dataset. Note that the environmental chamber had a temperature excursion for cycles 50–60.
**(a)** Voltage vs. discharge capacity for the first 100 cycles. The curves shift only subtly as a function of cycle number. Given the high discharge rates (4C), the voltage response is smoothed due to heterogeneity[38], and diagnostic differential capacity and/or differential voltage analysis cannot be applied.
**(b)** Capacity matrix representation of the data presented in **(a)**. The color represents capacity as a function of voltage and cycle number.
**(c)** The capacity matrix presented in **(b)**, subtracted by the discharge capacity of cycle 2 ("baseline-subtracted capacity matrix"). The small differences between cycles are now clearer.
**(d)** $Q_n - Q_2$ at two selected voltages as a function of cycle number.
**(e)** The capacity matrix presented in **(b)**, divided by the discharge capacity of cycle 2 ("baseline-divided capacity matrix"). The small differences between cycles are now clearer.
**(f)** $Q_n/Q_2$ at two selected voltages as a function of cycle number.

We also explore the "baseline-divided capacity matrix", $Q_n/Q_2$, which is the matrix presented in Figure 1b divided by the discharge capacity of cycle 2. Again, the small differences between cycles is now clearer. The region of maximum contrast occurs at a voltage ~300 mV higher for $Q_n/Q_2$ than $Q_n - Q_2$. Figure 1f presents $Q_n/Q_2$ at two selected voltages as a function of cycle number. The trends are linear here as well, as they were in Figure 1d. However, the trends are generally noisier and lower contrast in $Q_n/Q_2$ than in $Q_n - Q_2$. Additionally, $Q_n - Q_2$ is more physically meaningful than $Q_n/Q_2$, as discussed in the following paragraph. For these reasons, we focus on feature extraction from the baseline-subtracted capacity matrix, instead of the baseline-divided capacity matrix, in the remainder of this work.

Capacity matrices are clean, compact, machine-learning-ready representations of capacity vs. voltage and cycle number. Furthermore, baseline subtraction or division helps magnify the changes in the electrochemical response with cycle number due to degradation. Note that the sum of all elements in a column of the baseline-subtracted capacity matrix is proportional to the change in discharge energy between the constant-current portions of the cycle of interest and the baseline cycle (i.e., the integral of capacity over voltage); in fact, a single element is proportional to the change in discharge energy between the cycle of interest and the baseline cycle at a given voltage. While not strictly a requirement, the capacity matrix concept applies most naturally if some part of the cycling protocol is consistent with cycle number; capacity matrices could be developed with the charge data in this work as well, but in this case the electrochemical response would be convoluted by the interaction between the intrinsic degradation and the charging protocol. Furthermore, while we focus on the constant-current portion of the cycling protocol in this work, a similar concept could be applied to voltage vs. time during a constant-voltage hold, an open-circuit rest, or any step that is repeated during cycling. This approach may be especially interesting for lithium plating detection during voltage relaxation.[39–42]

## Feature generation and statistical learning approach

We now can represent all electrochemical discharge data for the first 100 cycles of each cell via a single capacity matrix with high dimensionality (1000 voltages × 100 cycles). In a sense, these matrices resemble single-channel (i.e., grayscale) images. One seemingly natural method to apply to this data is neural networks, which perform automatic feature learning and excel at high-dimensional data like images.[11] Indeed, these methods have been previously applied to this dataset.[8,24–28] However, neural networks typically require a large number of samples, are computationally expensive, require deep learning domain expertise, and—most importantly for our purposes—are often difficult to interpret. Our interest in this work is balancing accuracy with interpretability. Given the high performance of simple models in Severson et al.[3], we begin by exploring classical statistical learning methods here and consider deep learning approaches at a later point.

Overall, we maintain consistent modeling objectives as Severson et al.[3] Identically to Severson et al.[3], our objective is to predict the $\log_{10}$-transformed cycle life. We evaluate model performance with root-mean-squared error (RMSE), one of two performance metrics used by Severson et al.[3] While reducing the number of cycles is of interest, we also maintain the 100-cycle limit here to compare with previous results; Figure 5 of Severson et al.[3] demonstrates that similar predictive performance can be achieved with a similar modeling approach that uses only the first 60 cycles for prediction. Finally, we maintain the same training set and primary/secondary test sets as Severson et al.[3]; note that we exclude the outlier battery from the primary test set throughout this work.

A key step in statistical learning is feature generation—that is, proposing meaningful representations of the input data that predict the objective. In this case, where the number of features (100,000) greatly exceeds the number of observations (41), an associated objective is to reduce the dimensionality of a dataset to its "intrinsic dimensionality".[17] Here, many of the elements of the input data are highly correlated, i.e., data at cycle 50 is correlated with data at cycle 51, and data at 3.00 V is correlated with data at 3.01 V. Dimensionality reduction can be performed manually, by discarding some elements of the input data, or via dimensionality reduction methods that find alternative representations of the input data. Throughout this work, we consider both approaches.

One dimension that is straightforward to reduce is the number of voltage points used per discharge curve, as the choice of 1000 points per discharge curve in Severson et al.[3] was arbitrary. Figure 2 illustrates the effects of downsampling the number of points used to represent $\Delta Q_{100-10}(V)$. Figure 2a presents $\Delta Q_{100-10}(V)$ sampled with a different number of points for an example cell. Given the relatively featureless trends in the data, the key trend is captured with as few as 20 points. Figure 2b presents the RMSE of the univariate $\log_{10}(\mathrm{var}(\Delta Q_{100-10}(V)))$ model vs. the number of points used to evaluate the spline. These RMSEs are presented relative to the RMSE of the model using all 1000 points. Here, a new model is trained on the training data sampled at each sampling frequency and then evaluated on the two test sets. The change in the absolute value of RMSE exceeds 1% only after the sampling frequency exceeds 40 mV (40 points over a 1.6 V voltage window) for the training and primary test sets, and 160 mV (10 points) for the secondary test set. Clearly, 1000 points are not needed to capture the key features present in $\Delta Q_{100-10}(V)$ in this dataset. As such, we often reduce the number of points used to sample $\Delta Q_{100-10}(V)$ in the remainder of this work.

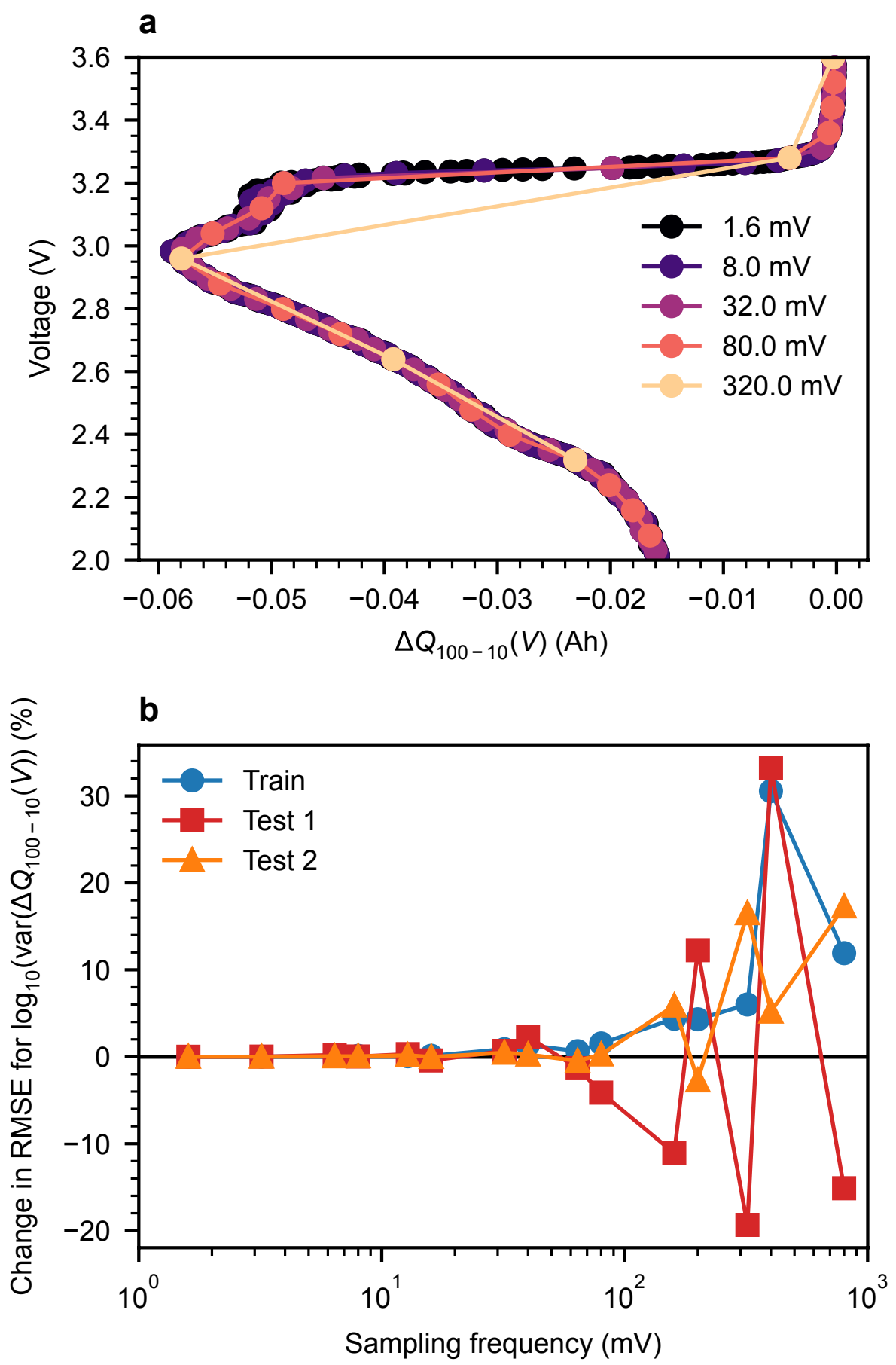

**Figure 2.** Dimensionality reduction via decreasing the number of points used to represent a single voltage-discharge capacity curve.

**(a)** Voltage vs. $\Delta Q_{100-10}(V)$ for an example cell as a function of the sampling frequency of the voltage points. The essential shape of the curve is maintained even with a sampling frequency of 80 mV, which is 50x less frequent than the 1.6 mV spacing used in Severson et al.[3]

**(b)** Change in RMSE of the univariate $\log_{10}(\mathrm{var}(\Delta Q_{100-10}(V)))$ model vs. the number of points used to evaluate the spline for the training, primary test, and secondary test sets. These RMSEs are expressed relative to the RMSE using default sampling frequency (1.6 mV, 1000 points/curve). Here, new models are trained on the training set, with $\Delta Q_{100-10}(V)$ data sampled at each sampling frequency. The change in the absolute value of RMSE exceeds 1% only after the sampling frequency exceeds 40 mV (40 points) for the training and primary test sets and 160 mV (10 points) for the secondary test set. Note that this result is likely sensitive to the fact that the discharge curve is applied at 4C; constant-current data at lower rates may require a higher sampling frequency to capture the essential shape of the curve.

## Univariate models from $\Delta Q_{100-10}(V)$

***Summary statistic-transformation pairings.***—Severson et al.[3] identified the $\log_{10}$ variance of $\Delta Q_{100-10}(V)$ as the highest-performing feature considered among the twenty features generated from all data sources. While Severson et al.[3] proposed plausible physical explanations, this feature remains somewhat unsatisfying from a battery science perspective. Specifically, the variance function is not directly physically meaningful when applied to the elements in $\Delta Q_{100-10}(V)$, though it may be correlated with more physically meaningful metrics such as the non-uniformity in the energy dissipation. Furthermore, the variance function ignores the ordering of the voltage points within $\Delta Q_{100-10}(V)$, which discards potentially useful information. In other words, summary statistics like variance treat all capacity points as unordered members of a population, while in fact they are sequentially ordered by voltage. An open question is if the $\log_{10}$ variance model indeed best captures the degradation trend in this dataset, or if another univariate transformation-function pairing can achieve similar results.

Here, we focus on univariate linear models with the intention of finding the simplest yet highest-performing models. We consider fourteen summary statistic functions of $\Delta Q_{100-10}(V)$ and four feature transformations, for a total of 56 function-transformation pairings. Most of these functions are summary statistics that are often applied to populations, e.g., mean, range, and variance; these functions map the $\Delta Q_{100-10}(V)$ vectors to scalars. We also explore feature transformation, which is a common preprocessing step that can improve model performance by improving the linearity of the relationship between the feature(s) and the output. The transformations considered here include no transformation, square root, cube root, and $\log_{10}$. However, one disadvantage of the square root and $\log_{10}$ transformations is that they do not accept negative values as input. Thus, we apply the absolute value function to summary statistics with both positively and negatively signed values for these two transformations. An obvious drawback of this approach is that values with equivalent magnitude but opposite signs are treated identically; we did not extensively explore alternative approaches such as power transforms[43]. Additionally, we use elastic net regression instead of ordinary least squares regression, even for these univariate models, so that the model coefficients can be regularized (i.e., given lower magnitude) in the interest of building more robust and generalizable models. Finally, we limit our analysis to leveraging cycles 100 and 10, i.e., $\Delta Q_{100-10}(V)$, and do not investigate other cycle number combinations. Figure 5 of Severson et al.[3] demonstrated that the accuracy of the cycle number combinations used in $\Delta Q_{100-10}(V)$ is relatively insensitive within this regime, but we acknowledge that other combinations of cycle numbers may outperform our choice here.

Figure 3 presents the performance of these 56 univariate models derived from these feature-transformation pairings to $\Delta Q_{100-10}(V)$ and fit via the elastic net. In this figure, each row represents a summary statistic, each column represents a transformation, and each panel represents a dataset (training, primary test, and secondary test, respectively). Here, IDR and IQR represent interdecile range (90$^{th}$ percentile–10$^{th}$ percentile) and interquartile range (75$^{th}$ percentile –25$^{th}$ percentile), respectively; for instance, the IQR of $\Delta Q_{100-10}(V)$ represents the difference between the 75$^{th}$ and 25$^{th}$ percentiles of the capacity values in the $\Delta Q_{100-10}(V)$ vector. The $\log_{10}$(IQR) model generally has the lowest error across all three data sets; its error is lower than that of the $\log_{10}$(var) model by 5, 14, and 6 cycles (4.8%, 10%, and 3.2%) for the training, primary test, and secondary test sets, respectively. Note that the test errors are consistently higher than the training error, which is consistent with Severson et al.[3] and possibly reflects differences between the datasets (e.g., the median cycle life of the secondary test set, 964.5 cycles, is 66% higher than that of the primary test set, 580 cycles).[3] All models requiring the use of the absolute value function perform poorly. Interestingly, simply evaluating $\Delta Q_{100-10}(V)$ at $V = 2.959$ V performs similarly to the IQR and variance models (the errors are 8.7%, −13.8%, and 7.1% higher than the variance model for the training, primary test, and secondary test sets, respectively). Note that this particular voltage was selected because the minimum of the baseline-subtracted capacity matrix occurs at this voltage (Figure 1c).

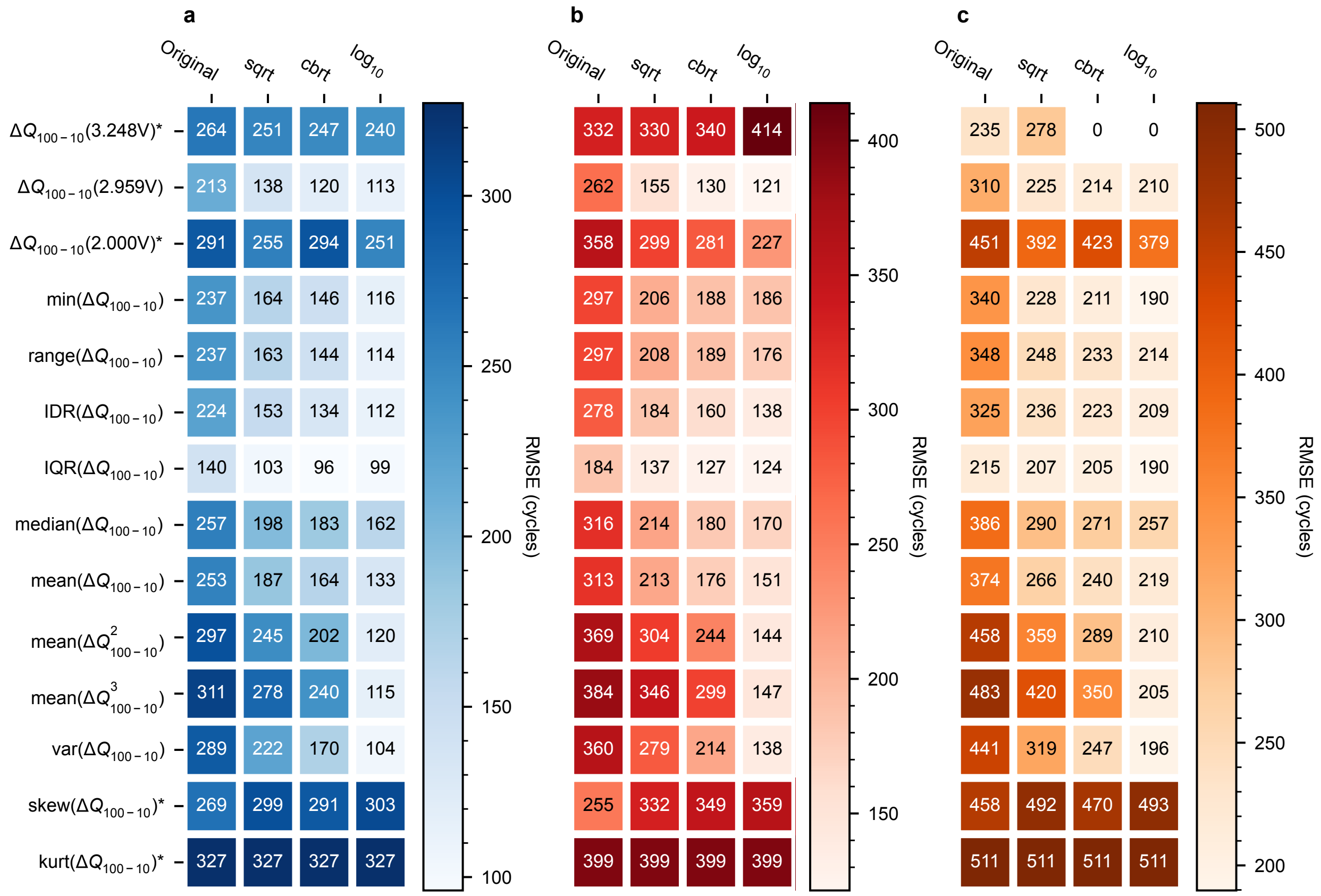

| (a) | Original | sqrt | cbrt | log10 |
|---|---|---|---|---|
| ΔQ100−10(3.248V)* | 264 | 251 | 247 | 240 |
| ΔQ100−10(2.959V) | 213 | 138 | 120 | 113 |
| ΔQ100−10(2.000V)* | 291 | 255 | 294 | 251 |
| min(ΔQ100−10) | 237 | 164 | 146 | 116 |
| range(ΔQ100−10) | 237 | 163 | 144 | 114 |
| IDR(ΔQ100−10) | 224 | 153 | 134 | 112 |
| IQR(ΔQ100−10) | 140 | 103 | 96 | 99 |
| median(ΔQ100−10) | 257 | 198 | 183 | 162 |
| mean(ΔQ100−10) | 253 | 187 | 164 | 133 |
| mean(ΔQ²100−10) | 297 | 245 | 202 | 120 |
| mean(ΔQ³100−10) | 311 | 278 | 240 | 115 |
| var(ΔQ100−10) | 289 | 222 | 170 | 104 |
| skew(ΔQ100−10)* | 269 | 299 | 291 | 303 |
| kurt(ΔQ100−10)* | 327 | 327 | 327 | 327 |

| (b) | Original | sqrt | cbrt | log10 |
|---|---|---|---|---|
| ΔQ100−10(3.248V)* | 332 | 330 | 340 | 414 |
| ΔQ100−10(2.959V) | 262 | 155 | 130 | 121 |
| ΔQ100−10(2.000V)* | 358 | 299 | 281 | 227 |
| min(ΔQ100−10) | 297 | 206 | 188 | 186 |
| range(ΔQ100−10) | 297 | 208 | 189 | 176 |
| IDR(ΔQ100−10) | 278 | 184 | 160 | 138 |
| IQR(ΔQ100−10) | 184 | 137 | 127 | 124 |
| median(ΔQ100−10) | 316 | 214 | 180 | 170 |
| mean(ΔQ100−10) | 313 | 213 | 176 | 151 |
| mean(ΔQ²100−10) | 369 | 304 | 244 | 144 |
| mean(ΔQ³100−10) | 384 | 346 | 299 | 147 |
| var(ΔQ100−10) | 360 | 279 | 214 | 138 |
| skew(ΔQ100−10)* | 255 | 332 | 349 | 359 |
| kurt(ΔQ100−10)* | 399 | 399 | 399 | 399 |

| (c) | Original | sqrt | cbrt | log10 |
|---|---|---|---|---|
| ΔQ100−10(3.248V)* | 235 | 278 | 0 | 0 |
| ΔQ100−10(2.959V) | 310 | 225 | 214 | 210 |
| ΔQ100−10(2.000V)* | 451 | 392 | 423 | 379 |
| min(ΔQ100−10) | 340 | 228 | 211 | 190 |
| range(ΔQ100−10) | 348 | 248 | 233 | 214 |
| IDR(ΔQ100−10) | 325 | 236 | 223 | 209 |
| IQR(ΔQ100−10) | 215 | 207 | 205 | 190 |
| median(ΔQ100−10) | 386 | 290 | 271 | 257 |
| mean(ΔQ100−10) | 374 | 266 | 240 | 219 |
| mean(ΔQ²100−10) | 458 | 359 | 289 | 210 |
| mean(ΔQ³100−10) | 483 | 420 | 350 | 205 |
| var(ΔQ100−10) | 441 | 319 | 247 | 196 |
| skew(ΔQ100−10)* | 458 | 492 | 470 | 493 |
| kurt(ΔQ100−10)* | 511 | 511 | 511 | 511 |

**Figure 3.** Root-mean-square error (RMSE) of univariate models derived from $\Delta Q_{100-10}(V)$ as a function of summary statistic (rows) and transformation (columns) for the **(a)** training set, **(b)** primary test set, and **(c)** secondary test set in predicting the $\log_{10}$-transformed cycle life. IDR and IQR represent interdecile range and interquartile range, respectively. The $\log_{10}$(IQR) model generally has the lowest error across all three data sets, even exceeding the $\log_{10}$(var) model. The asterisk for some summary statistics denotes models where the absolute value was applied before the log/square root transformation to ensure positive values were used as input for these transformations. Models with anomalously high error are excluded from these plots and indicated using “0”.

Generally, low error for a linear model signifies an unbiased linear relationship between the transformed feature and the prediction target. We first note the effectiveness of the $\log_{10}$ transformation, which uniformly leads to the lowest errors of the four transformations considered. The $\log_{10}$ transformation is considered among the most powerful in reducing right skew[17] though should be used with caution[44]; because the distribution of both features and cycle lives is right skewed (i.e., many more cells have low cycle life than high cycle life) in the training set, the $\log_{10}$ transformation helps linearize a trend that would otherwise be skewed by the most extreme values in the training set. Furthermore, the $\log_{10}$ transformation may reduce error simply because the true relationship between the features and the response is logarithmic. We return to this point in our discussion of multivariate models.

The highest performing functions are variance (identified by Severson et al.[3]), IQR, IDR, and $\Delta Q_{100-10}(V)$ at $V = 2.959$ V. One commonality between the variance, IQR, and IDR functions is that all are measures of dispersion, or spread. Here, higher dispersion among the elements of $\Delta Q_{100-10}(V)$ correlates with poor cycle life; naturally, cells with higher dispersion have larger changes from cycle to cycle and thus lower cycle life. However, the reasons for the higher performance of measures of dispersion over measures of central tendency (e.g., mean, median) is not immediately obvious. We revisit this aspect throughout the rest of this work.

***Cycle life transforming and cycle averaging.***—Next, we consider two additional variations of Figure 3: using the observed cycle life instead of the $\log_{10}$-transformed cycle life (Figure 4) and local cycle averaging (Figure 5). First, we study the importance of transforming the prediction objective. Figure 4 is equivalent to Figure 3 but uses the observed cycle life instead of the $\log_{10}$- transformed cycle life as the prediction objective. The errors using observed cycle life are consistently higher than those using the $\log_{10}$-transformed cycle life, demonstrating the benefits of log transforming the prediction objective for this long-tailed (i.e., right-skewed) dataset.

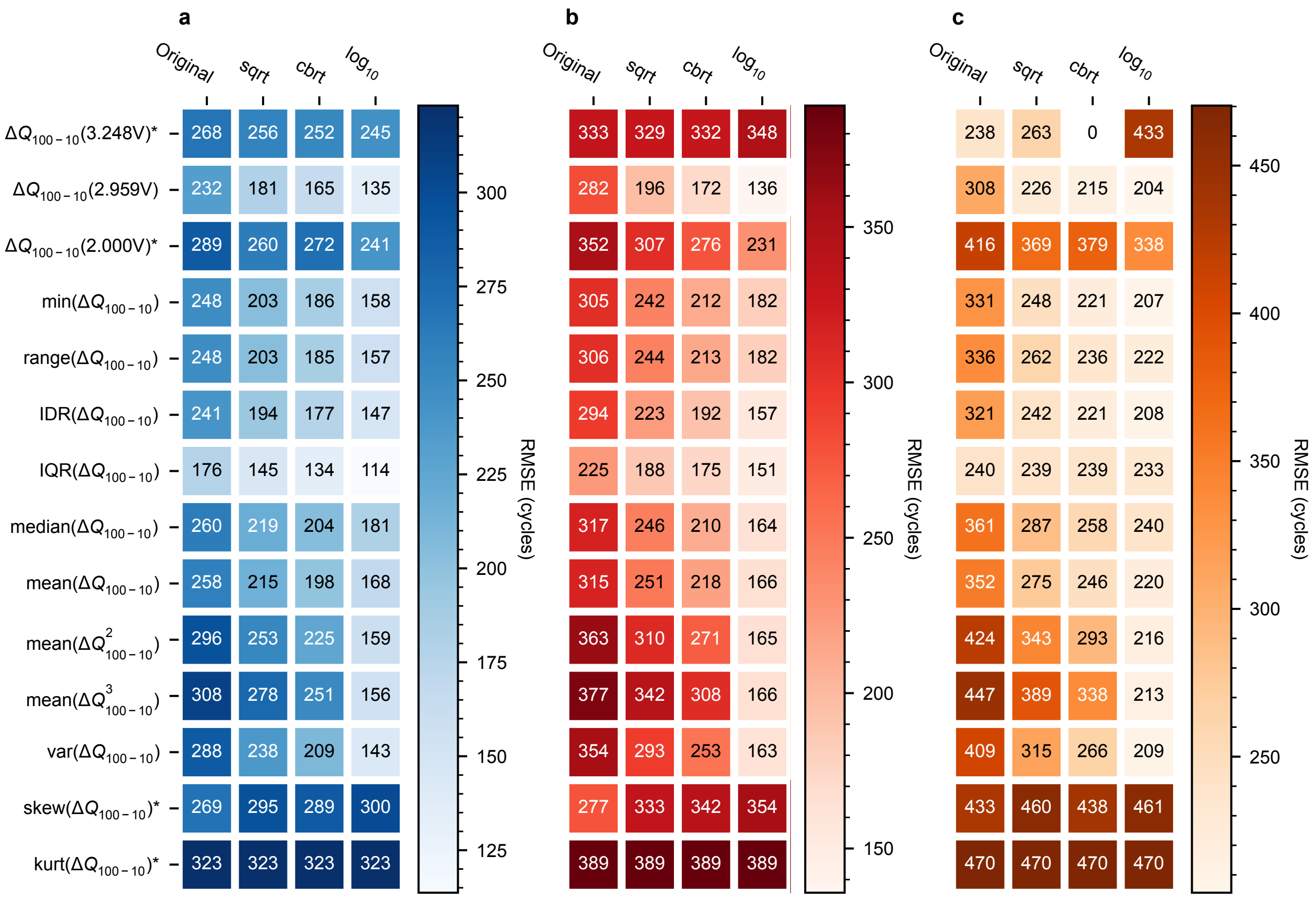

**Figure 4.** Root-mean-square error (RMSE) of univariate models derived from $\Delta Q_{100-10}(V)$ as a function of summary statistic (rows) and transformation (columns) for the **(a)** training set, **(b)** primary test set, and **(c)** secondary test set in predicting cycle life (as opposed to the $\log_{10}$-transformed cycle life, as in Figure 3). IDR and IQR represent interdecile range and interquartile range, respectively. The $\log_{10}$(IQR) model generally has the lowest error across all three data sets, even exceeding the $\log_{10}$(var) model. However, the errors of these models are uniformly higher than the errors of corresponding models predicting the $\log_{10}$-transformed cycle life. The asterisk for some summary statistics denotes models where the absolute value was applied before the log/square root transformation to ensure positive values were used as input for these transformations. Models with anomalously high error are excluded from these plots and indicated using “0”.

Second, Figure 5 presents local cycle averaging, denoted $\Delta Q_{98:100-9:11}(V)$; here, the "98:100" nomenclature denotes inclusively averaging these cycles across voltages. The motivation for cycle averaging was to improve the signal-to-noise ratio inherent to using only one cycle. However, the performance of these models is generally comparable with or without cycle averaging, generally varying by a few cycles in either direction. Thus, the signal-to-noise ratio does not appear to improve with local cycle averaging.

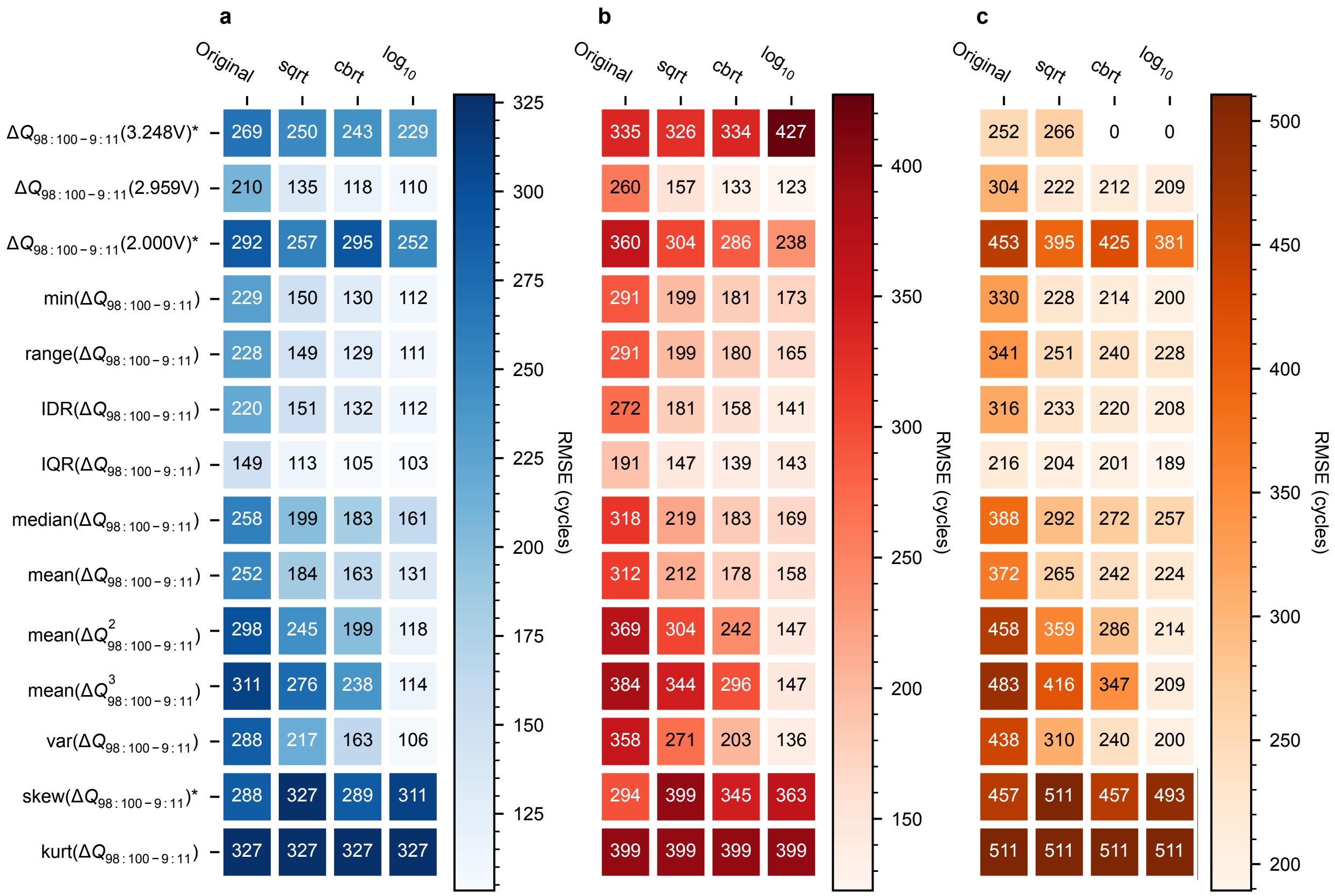

| | a Original | a sqrt | a cbrt | a log10 | b Original | b sqrt | b cbrt | b log10 | c Original | c sqrt | c cbrt | c log10 |
|---|---|---|---|---|---|---|---|---|---|---|---|---|
| $\Delta Q_{98:100-9:11}(3.248V)$* | 269 | 250 | 243 | 229 | 335 | 326 | 334 | 427 | 252 | 266 | 0 | 0 |
| $\Delta Q_{98:100-9:11}(2.959V)$ | 210 | 135 | 118 | 110 | 260 | 157 | 133 | 123 | 304 | 222 | 212 | 209 |
| $\Delta Q_{98:100-9:11}(2.000V)$* | 292 | 257 | 295 | 252 | 360 | 304 | 286 | 238 | 453 | 395 | 425 | 381 |
| $\min(\Delta Q_{98:100-9:11})$ | 229 | 150 | 130 | 112 | 291 | 199 | 181 | 173 | 330 | 228 | 214 | 200 |
| $\mathrm{range}(\Delta Q_{98:100-9:11})$ | 228 | 149 | 129 | 111 | 291 | 199 | 180 | 165 | 341 | 251 | 240 | 228 |
| $\mathrm{IDR}(\Delta Q_{98:100-9:11})$ | 220 | 151 | 132 | 112 | 272 | 181 | 158 | 141 | 316 | 233 | 220 | 208 |
| $\mathrm{IQR}(\Delta Q_{98:100-9:11})$ | 149 | 113 | 105 | 103 | 191 | 147 | 139 | 143 | 216 | 204 | 201 | 189 |
| $\mathrm{median}(\Delta Q_{98:100-9:11})$ | 258 | 199 | 183 | 161 | 318 | 219 | 183 | 169 | 388 | 292 | 272 | 257 |
| $\mathrm{mean}(\Delta Q_{98:100-9:11})$ | 252 | 184 | 163 | 131 | 312 | 212 | 178 | 158 | 372 | 265 | 242 | 224 |
| $\mathrm{mean}(\Delta Q^2_{98:100-9:11})$ | 298 | 245 | 199 | 118 | 369 | 304 | 242 | 147 | 458 | 359 | 286 | 214 |
| $\mathrm{mean}(\Delta Q^3_{98:100-9:11})$ | 311 | 276 | 238 | 114 | 384 | 344 | 296 | 147 | 483 | 416 | 347 | 209 |
| $\mathrm{var}(\Delta Q_{98:100-9:11})$ | 288 | 217 | 163 | 106 | 358 | 271 | 203 | 136 | 438 | 310 | 240 | 200 |
| $\mathrm{skew}(\Delta Q_{98:100-9:11})$* | 288 | 327 | 289 | 311 | 294 | 399 | 345 | 363 | 457 | 511 | 457 | 493 |
| $\mathrm{kurt}(\Delta Q_{98:100-9:11})$ | 327 | 327 | 327 | 327 | 399 | 399 | 399 | 399 | 511 | 511 | 511 | 511 |

**Figure 5.** Root-mean-square error (RMSE) of univariate models derived from $\Delta Q_{98:100-9:11}(V)$ as a function of summary statistic (rows) and transformation (columns) for the **(a)** training set, **(b)** primary test set, and **(c)** secondary test set in predicting the $\log_{10}$-transformed cycle life. The "98:100" syntax in the subscript denotes inclusive averaging of these cycles across voltage. IDR and IQR represent interdecile range and interquartile range, respectively. The $\log_{10}$(IQR) model generally has the lowest error across all three data sets, even exceeding the $\log_{10}$(var) model. However, the errors of these models are generally comparable to the errors of corresponding

models that use $\Delta Q_{100-10}(V)$ directly (i.e., without cycle averaging). The asterisk for some summary statistics denotes models where the absolute value was applied before the log/square root transformation to ensure positive values were used as input for these transformations. Models with anomalously high error are excluded from these plots and indicated using "0".

***Univariate percentile models.***—Given the success of the univariate IQR and IDR models, we explored models using other percentile ranges of $\Delta Q_{100-10}(V)$. Figure 6 presents univariate models based on the $\log_{10}$ of different percentile ranges of $\Delta Q_{100-10}(V)$. In Figures 6a–6c, the errors of models using various upper and lower percentiles (swept in 1% intervals) are presented for the training, primary test, and secondary test sets, respectively. In these plots, the diagonal elements (i.e., equal lower and upper percentiles) represent a specific percentile of $\Delta Q_{100-10}(V)$, e.g., the 50$^{th}$ percentile, as opposed to the percentile difference as in non-diagonal elements, e.g., the 75$^{th}$ percentile minus the 25$^{th}$ percentile. The errors of models trained on each of the datasets are generally minimized for lower percentile values between 30–60% and for upper percentile values between 40–80%: the training set error is minimized for a lower percentile of 31% and an upper percentile of 62%, the primary test set error is minimized for a lower percentile of 43% and an upper percentile of 64%, and the secondary test set error is minimized for a lower percentile of 1% and an upper percentile of 75%. Using only the model trained on the training set (i.e., using the 31$^{st}$ and 62$^{nd}$ percentiles), we find that while the training and primary test set errors are quite low (82 and 109 cycles, respectively), the secondary test set error is high (261 cycles). Note that the minimum error of the secondary test set is nearly 100 cycles lower (169 cycles) using the 1–75% percentile range than using the 31–62% percentile range. Interestingly, the primary test set error is lower with the percentile model than the IQR model (109 vs. 124 cycles), suggesting the primary and secondary test sets may have some inherent differences (e.g., the small positive values of $\Delta Q_{100-10}(V)$ at around 3.25 V in the secondary test set). Note that the training and primary test sets consist of cells from the same two cycling batches (2017-05-12 and 2017-06-30), while the secondary test set consists of cells from a distinct cycling batch (2018-04-12) that was cycled nearly a year later, which led to an additional ~0.01 Ah of calendar aging, or ~1% of the nominal capacity (Supplementary Table 2 of Attia et al.[22]). Since cycle life is defined as 80% of nominal capacity (i.e., 0.88 Ah), the maximum capacity loss over useful life is 20% of nominal capacity (i.e., 0.22 Ah); thus, an initial capacity decrease of 1% of nominal capacity (i.e., 0.011 Ah) corresponds to a 5% decrease in the maximum capacity loss available over useful life (i.e., 0.011 Ah / 0.22 Ah). Another interesting comparison is to the IQR model of Figure 3: the training set error of the 31–62% percentile model is lower (82 cycles vs. 99 cycles) but the secondary test set error is significantly higher (261 cycles vs. 190 cycles). Overall, the percentile model may be overfitting to the training set and thus more poorly generalizing to the somewhat different secondary test set.

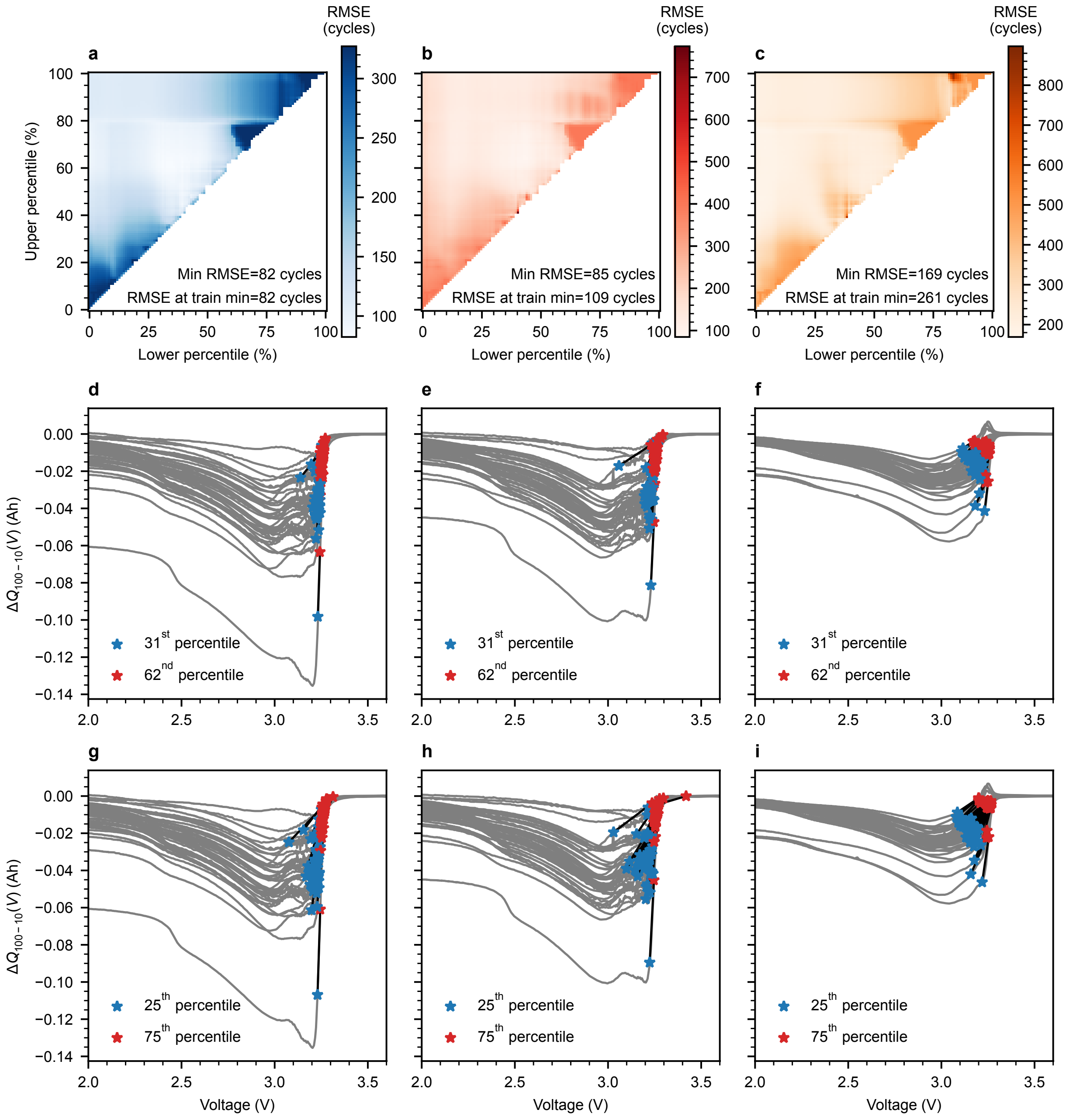

**Figure 6.** Univariate models based on the $\log_{10}$ of the percentile difference of $\Delta Q_{100-10}(V)$. The RMSEs of these models as a function of the lower and upper percentile used in the percentile difference are displayed for the **(a)** training set, **(b)** primary test set, and **(c)** secondary test set. In these plots, the diagonal elements (i.e., equal lower and upper percentiles) represent the given percentile of $\Delta Q_{100-10}(V)$, as opposed to the percentile difference as in non-diagonal elements (note that the diagonal elements generally have higher error than the non-diagonal elements, meaning that the percentile differences outperform the percentiles). Additionally, $\Delta Q_{100-10}(V)$ and its $31^{st}$ and $62^{nd}$ percentiles are displayed for all cells in the **(d)** training set, **(e)** primary test

set, and **(f)** secondary test set; $\Delta Q_{100-10}(V)$ and its 25th and 75th percentiles (i.e., IQR) are displayed for all cells in the **(g)** training set, **(h)** primary test set, and **(i)** secondary test set.

Figures 6d–6f display the points representing the 31st and 62nd percentiles of $\Delta Q_{100-10}(V)$ for all cells in the training, primary test, and secondary test sets, respectively. Additionally, Figures 6g–6i display the points representing the 25th and 75th percentiles of $\Delta Q_{100-10}(V)$, or IQR($\Delta Q_{100-10}(V)$), for all cells in the training, primary test, and secondary test sets, respectively. For almost all cells in the training and primary test sets, both the upper and lower percentiles happen to be located on the sharp shoulder near 3.2 V, which corresponds to a change in the largest graphite plateau ($x_{graphite}$ = ~0.5 to ~1.0)[45]. This observation helps rationalize the success of the dispersion-based univariate models of Figure 3: in some sense, this feature is a measure of the length of the shoulder in $\Delta Q_{100-10}(V)$, which appears to be more predictive of cycle life than the mean or median of $\Delta Q_{100-10}(V)$. Note that both of these percentile ranges of $\Delta Q_{100-10}(V)$ generally do not capture the sharp shoulder at ~3.2 V for the secondary test set, which may explain why these models do not perform well on this dataset. Based on this result, an engineered feature that may perform well for all datasets is the magnitude of the shoulder at ~3.2 V, although we do not explore this type of curated feature engineering in this work.

***Univariate single-element models.***—Finally, we consider univariate models derived from single elements of $\Delta Q_{100-10}(V)$, i.e., evaluating the difference between two discharge capacity curves at a single voltage. Inspired by the success of the model using only $\Delta Q_{100-10}(V)$ at $V$ = 2.959 V, these models use only one element in the entire 100,000-element capacity matrix. Figure 7 displays results from univariate models derived from single elements of $\Delta Q_{100-10}(V)$, i.e., evaluated at a single voltage point; the errors of the training set, primary test set, and secondary test set are displayed in Figures 7a–7c, respectively. The same four transformations as before are considered: the original data, square root, cube root, and $\log_{10}$. These models are trained on the training set and evaluated on the primary and secondary test sets. Again, the errors of the $\log_{10}$ models are clearly the lowest out of these four transformations, although the errors from the cube root transformation are comparable. With the $\log_{10}$ transformation, the errors are minimized at 2.911 V for the training set (RMSE=108 cycles), 2.742 V for the primary test set (RMSE=109 cycles), and 2.998 V for the secondary test set (RMSE=204 cycles). The error of the primary test set is minimized at a voltage 256 mV lower than the voltage that minimizes the error of the secondary test set, although the RMSE of the $\log_{10}$ model evaluated on the primary test set is noisy within this voltage regime. The test set RMSEs evaluated at the voltage that minimizes the training set error for the $\log_{10}$-transformed models (2.911 V) are 118 cycles and 214 cycles for the primary and secondary test sets, respectively—errors that, remarkably, are only marginally higher (−4.8% and 12.6%) than the errors of the $\log_{10}$(IQR) model.

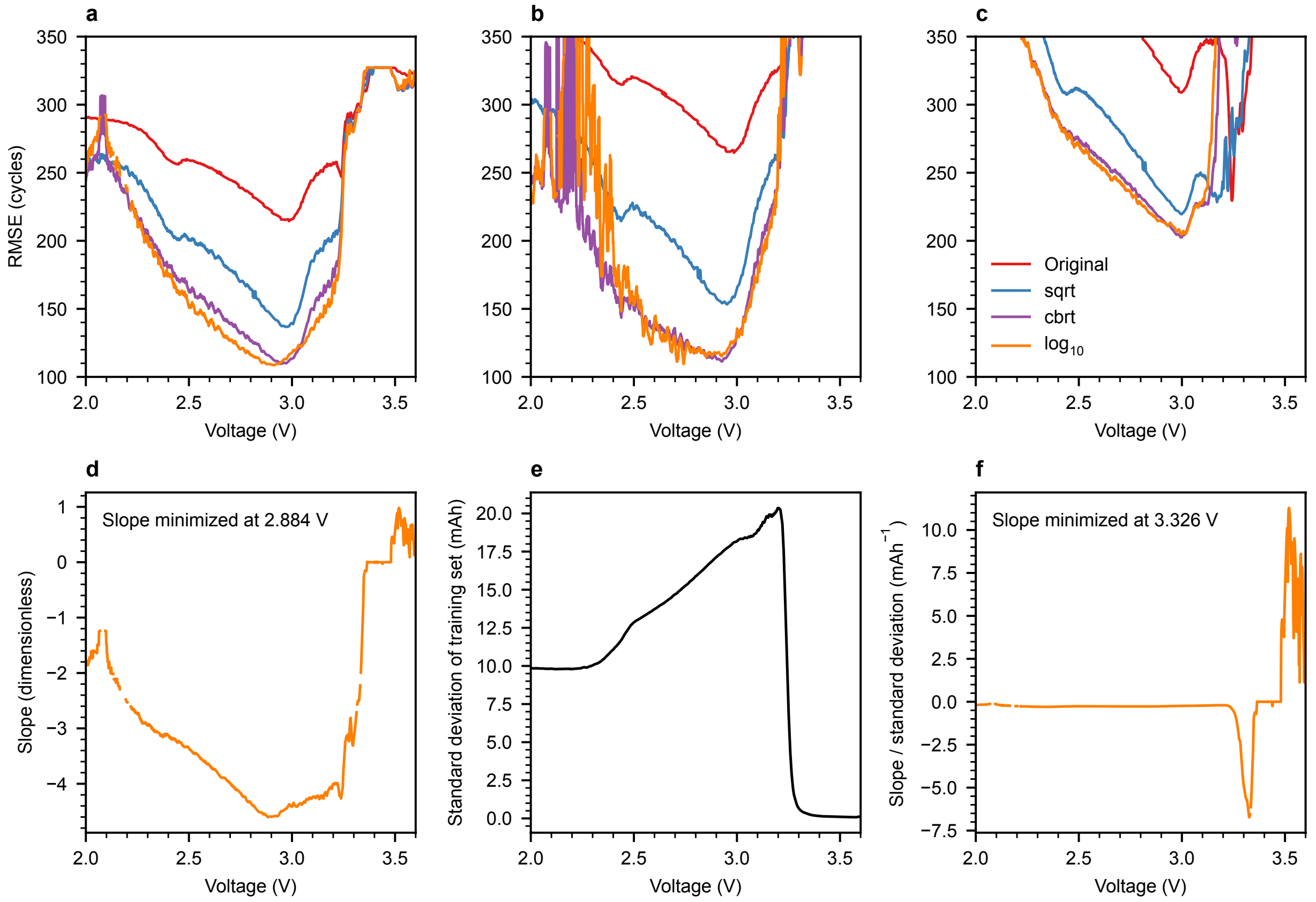

**Figure 7.** Univariate models derived from single elements of $\Delta Q_{100-10}(V)$, i.e., evaluated at a single voltage point. The RMSEs of these models with four transformations (original data, square root, cube root, and $\log_{10}$) are displayed for the **(a)** training set, **(b)** primary test set, and **(c)** secondary test set. The slopes (i.e., the coefficient of the linear model trained at a specific voltage) of the $\log_{10}$ models as a function of voltage are displayed in **(d)**; note that because the standardized features were used as input, the mean and standard deviation of each feature in the training set influences the slopes. The standard deviation of the features in the training set as a function of voltage is displayed in **(e)**, and the slopes from the $\log_{10}$ models divided by the standard deviation is displayed in **(f)**.

Figure 7d displays the slope of the linear models vs. the voltage. These slopes (i.e., the coefficient of the linear model trained at a specific voltage) are dimensionless since they use dimensionless input data (due to standardization) to predict the dimensionless $\log_{10}$-transformed cycle life. The slope is minimized at 2.884 V, suggesting that cycle life is most sensitive to $\Delta Q_{100-10}(V)$ at this voltage. However, this result is also influenced by the standard deviation of the training set as a function of voltage, which varies widely with voltage (Figure 7e). A fairer measure of voltage sensitivity, the slope divided by the standard deviation, is presented in Figure 7f. Here, we find that the voltage that produces the univariate model with the smallest standard-deviation-normalized slope is 3.326 V, a voltage that corresponds to the beginning of the large shoulder in $\Delta Q_{100-10}(V)$; at this voltage, the contrast between cells with low and high cycle life is high. This result is consistent with our results from Figure 6 that indicate the predictive nature of this shoulder. Note that the absolute value of this metric is maximized at voltages between ~3.5 V and ~3.6 V, where the standard deviation is very low, but these models have high RMSEs (Figure 7a–7c). Overall, the high performance of these simple univariate models highlights how easily interpretable models can still maintain high accuracy.

### Multivariate models from $\Delta Q_{100-10}(V)$

Thus far, we have primarily discussed univariate linear models of the form $y = mx + b$. While the results from these models are satisfactory for many applications (e.g., closed-loop optimization[22]), model performance typically improves as additional features are added—as long as these additional features capture meaningful information about the underlying prediction objective. In this section, we explore using the elements of $\Delta Q_{100-10}(V)$ directly as input features for model building. An important outcome of this approach is that the model coefficients can provide some clues as to which features (i.e., capacity differences at a given voltage) have the largest impact on cycle life (either positive or negative). However, a challenge with this approach is the high collinearity of the features, as $\Delta Q_{100-10}(V)$ at one voltage will be closely related to the value at a neighboring voltage. Fortunately, this situation is common in other fields like chemometrics and bioinformatics, and many statistical learning methods have been developed for these types of applications.

We consider four statistical learning methods and two nonlinear methods for this task, many of which are recommended for applications with many highly correlated features.[17] We briefly review them here. Ridge regression combines ordinary least squares (OLS) regression with a penalty on the sum of the square of the magnitude of the coefficients, which reduces the tendency to overfit to noise in the features.[17] Elastic net regression is similar to ridge regression but adds an additional penalty on the sum of the absolute value of the coefficients, which can cause the weights of some predictor values to be zero (i.e., not used in the final model).[17,31] Principal components regression (PCR) first performs principal components analysis on the features, the principal components of which maximize the explained variance of the features, and then performs linear regression on these principal components.[17] Similarly, partial least squares regression (PLSR), also known as projection to latent structures regression, uses transformations that maximize the variance of the features, but in this case the variance of the target variable (i.e., cycle life) is also used in building the components.[17,46] In other words, PLSR attempts to identify dimensions of the data with both high variance and a strong correlation with the response. Random forest regression aggregates the predictions of multiple decision trees training on a subset of the available observations and features in the training set.[17] Although random forest regression does not result in a linear model, this method has “feature importance” metrics that can aid in interpretability; Gini importance[17] is used in this work. Finally, multi-layer perceptrons (MLP) are feed-forward neural networks. The specific architectures of such models vary; here we use fully connected layers and rectified linear unit (ReLU) nonlinearities. Interpreting the MLP predictions is not necessarily straightforward; while various tools have been proposed to interpret otherwise black box models, we use SHapley Additive exPlanations (SHAP)[47] here.

Here, we used 100 features as input by downsampling the 1000-feature set (16 mV sampling frequency), as some methods (specifically, elastic net and random forest) did not converge during model fitting when using the full 1000-feature set as input. For all methods except MLP, the features were standardized during preprocessing, i.e., mean-subtracted and scaled by the standard deviation of the training set; for the MLP, the features were only scaled by the standard deviation of the training set. All models were trained via 5-fold cross validation. Note that we use the untransformed features here, as the square root and $\log_{10}$ transformations do not accept negative values as input. We also considered the cube root transformation, which is not subject to this limitation, but we generally obtained higher errors with this transformation.

Figure 8 presents the results of this approach using these methods. Figure 8a presents a bar plot of the training, primary test, and secondary test errors for each of the six methods. Interestingly, the nonlinear models, RF and MLP, have the worst test set performance, while ridge regression and PLSR have the best test set performance. Overall, the PLSR model generally has the lowest errors for all three datasets (59, 100, and 176 cycles for the training, primary test, and secondary test sets, respectively), and, in fact, has lower test error than any of the univariate models considered in this work (99, 124, and 190 cycles for the training, primary test, and secondary test sets, respectively for the IQR model). This model also has lower test error than the "full model" from Severson et al.[3] (51, 100, and 214 cycles for the training, primary test, and secondary test sets, respectively) and comparable test error to the "discharge model" from Severson et al.[3] (76, 86, and 173 cycles for the training, primary test, and secondary test sets, respectively). Both of these models from Severson et al.[3] use features from additional sources, while the PLSR model uses only $\Delta Q_{100-10}(V)$ as input. Figure 8b is identical to Figure 8a but uses $\Delta Q_{98:100-9:11}(V)$ (i.e., cycle-averaged) instead of $\Delta Q_{100-10}(V)$. The results in Figure 8a and 8b are comparable, suggesting cycle averaging does not help in this case.

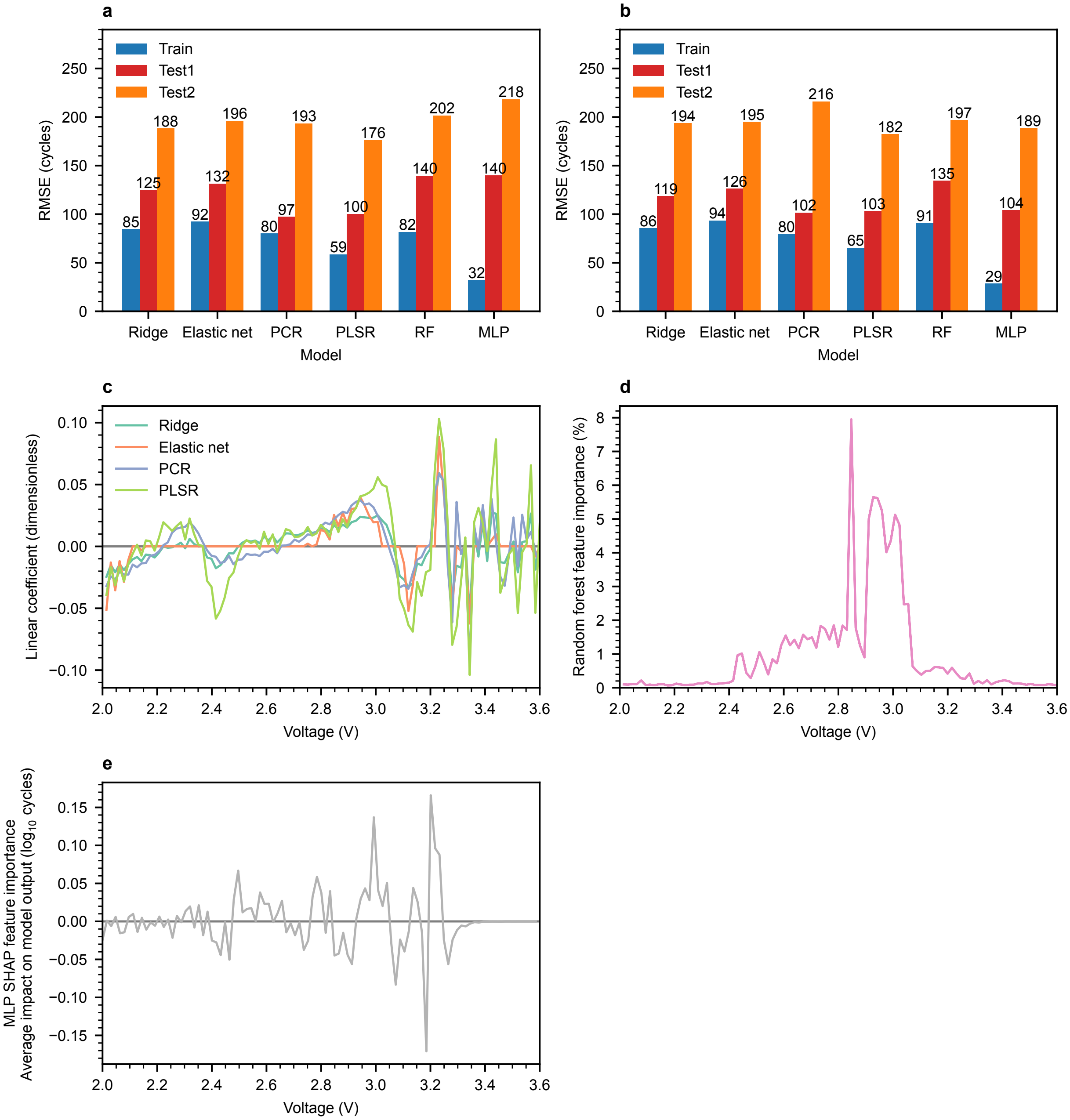

**Figure 8.** Multivariate models built via various methods using the elements of $\Delta Q_{100-10}(V)$ as input features. Here, PCR represents principal component analysis, PLSR represents partial least

square regression (also known as projection to latent structures regression), RF represents random forest regression, and MLP represents multi-layer perceptron regression.
**(a)** RMSE for the training set, primary test set, and secondary test set for each of the five methods using $\Delta Q_{100-10}(V)$. PLSR generally has the lowest errors.
**(b)** RMSE for the training set, primary test set, and secondary test set for each of the five methods using $\Delta Q_{98:100-9:11}(V)$. PLSR generally has the lowest errors. The errors are comparable to the non-cycle averaged case.
**(c)** Scaled coefficients of the linear models to predict $\log_{10}$ cycle life for the four methods that produce linear models. These coefficients are generally consistent across all four methods.
**(d)** Random forest feature importance (Gini importance) vs. voltage. The most important features are somewhat different from those indicated by the linear models, with little emphasis on the shoulder at ~3.2 V.
**(e)** MLP feature importance (SHAP importance) vs. voltage. The general trends are similar to the linear models.

Figure 8c displays the feature coefficients for the four methods that produce linear models (ridge regression, elastic net regression, PCR, and PLSR). Because $\Delta Q_{100-10}(V)$ is generally negative, negative coefficients correspond to an increase in cycle life (i.e., the negative feature in $\Delta Q_{100-10}(V)$ multiplied by a negative coefficient is positive). Note that these coefficients do not directly correspond to the features that have the largest impact on cycle life because the standardized features were used as input, and thus these coefficients depend on both the relationship with cycle life and the mean and standard deviation of the raw features. We illustrate this point later in this discussion.

Overall, the four statistical learning methods produce remarkably consistent coefficient vectors, which provides confidence in these models. The coefficient vectors indicate that different voltage domains have clear positive and negative effects on cycle life, which is perhaps surprising since nearly all elements in $\Delta Q_{100-10}(V)$ are more negative for cells with lower cycle life (Figure 9a). Above 3.2 V, the linear coefficients change sign frequently and have high magnitude, although the high magnitude is likely due to the low standard deviation of the training features in this regime. The coefficients between ~3.05 V and ~3.2 V are consistently negative in all models (i.e., large negative values of $\Delta Q_{100-10}(V)$ increase cycle life), while the coefficients between ~2.5 V and ~3.05V are consistently positive (i.e., large negative values of $\Delta Q_{100-10}(V)$ decrease cycle life). We hypothesize that the frequently changing signs of the elements in the coefficient vectors are consistent with the high performance of dispersion-based univariate models trained on $\Delta Q_{100-10}(V)$, as opposed to the models based on central tendency metrics (e.g., mean and median). While dispersion-based univariate models also perform well, using the elements of $\Delta Q_{100-10}(V)$ directly as input features for multivariate models better captures the effects of this variation.

Figure 8d displays feature importance from the random forest model. The most important features in this model are somewhat different from those indicated by the statistical learning methods that produce linear models, with little emphasis on the shoulder at ~3.2 V; however, the specific values of voltages that are emphasized are not surprising in light of the results from the other models. Figure 8e displays the feature importance from the MLP model. The MLP has similarities to both the linear and random forest models; its most important features are at ~3.0 V and ~3.2 V. Overall, interpreting either of these methods is less straightforward than linear models; the development of tools for interpreting complex nonlinear models is an open area of research. These challenges illustrate the value of using inherently interpretable methods (e.g., methods producing linear models) over the explainability features of blackbox machine learning methods.[16]

Figure 9 illustrates interpretation of the PLSR coefficient vector via the use of examples. Figure 9a displays $\Delta Q_{100-10}(V)$ for all cells in the training set, with five selected cells colored by cycle life. The elements of $\Delta Q_{100-10}(V)$ are consistently more negative for cells with lower cycle life, although these curves do have some subtle differences as a function of voltage (e.g., the negative peak at 3.2 V for the cell with a cycle life of 300 cycles). Similarly, Figure 9b displays the standardized $\Delta Q_{100-10}(V)$ for all cells in the training set, again with five selected cells colored by cycle life. The standardized curves are noisy and not ordered by cycle life above ~3.3 V, suggesting that this data is not meaningful for predicting cycle life. However, below ~3.3 V, the standardized values of $\Delta Q_{100-10}(V)$ are fairly consistent for a given cell as a function of voltage, suggesting that the $\Delta Q_{100-10}(V)$ trends for most cells in the training set are simply scalar multiples of each other. Furthermore, note that the gap between both $\Delta Q_{100-10}(V)$ and standardized $\Delta Q_{100-10}(V)$ is much larger for the cells with cycle lives of 300 and 461 cycles (~4 standard deviations at 2.8 V) than the cells with cycle lives of 1424 and 2160 cycles (~0.5 standard deviations at 2.8 V), despite the fact that the difference in cycle number is small. This result helps rationalize our previous observation of the effectiveness of the $\log_{10}$ transformation for both the features and for cycle life, as each unit increase in the magnitude of $\Delta Q_{100-10}(V)$ has a diminishing effect on cycle life.

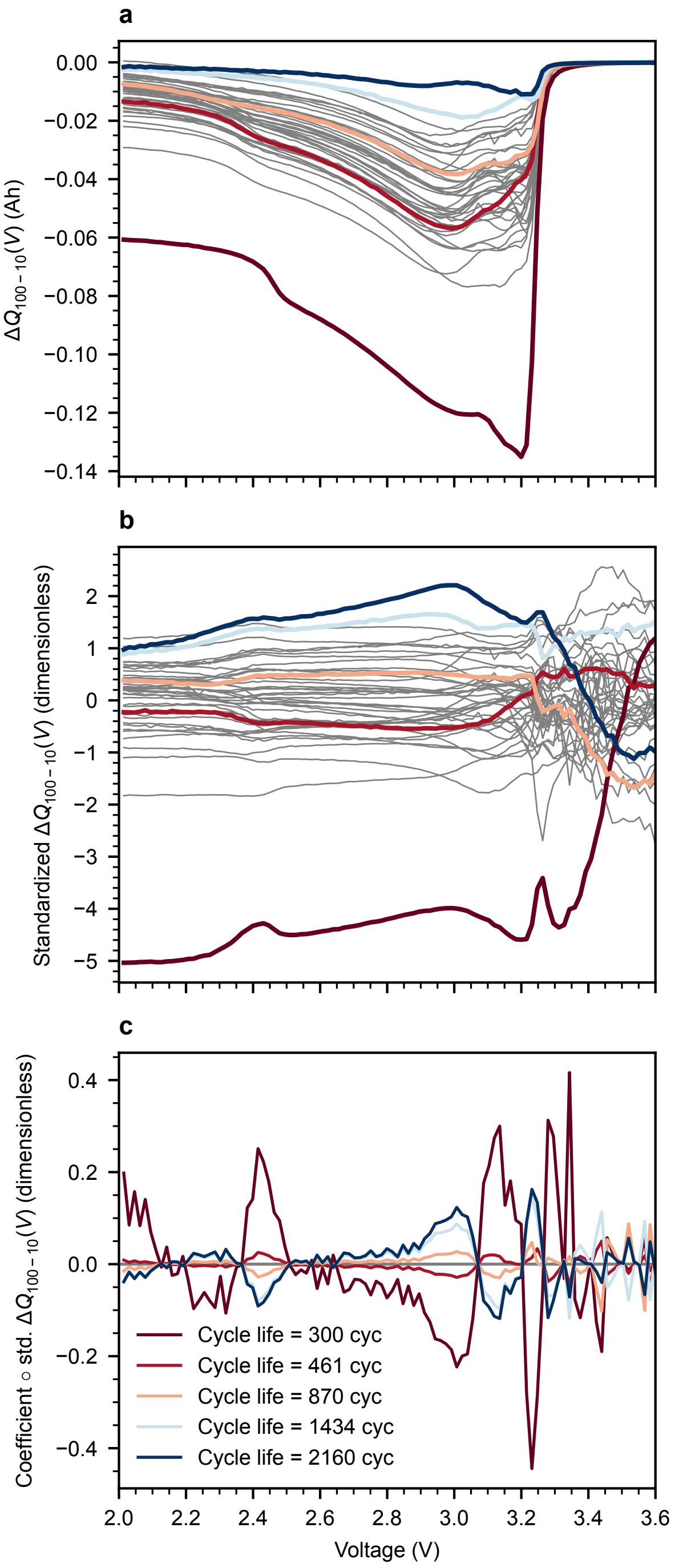

**Figure 9.** Illustration of the PLSR model for five example cells from the training set.
**(a)** $\Delta Q_{100-10}(V)$ for all cells in the training set, with five selected cells colored by cycle life. In general, the elements of $\Delta Q_{100-10}(V)$ are more negative for cells with lower cycle life. Cycle lives can be found in the legend of panel (c).
**(b)** Standardized $\Delta Q_{100-10}(V)$ for all cells in the training set, with the same five selected cells colored by cycle life.

**(c)** The element-wise product of the linear coefficient vector from the PLSR model and standardized $\Delta Q_{100-10}(V)$ for these example cells. In general, this product is dramatically different for the cells with extreme values of cycle life, and different voltage regions have different relationships with the response. Note that positive values are positively correlated with the $\log_{10}$-transformed cycle life.

Figure 9c displays the element-wise product of the linear coefficient vector from the PLSR model and $\Delta Q_{100-10}(V)$ for these example cells. The elements of this product are added together, along with the intercept, to calculate the $\log_{10}$-transformed cycle life (note that positive values are positively correlated with the $\log_{10}$-transformed cycle life). In general, this element-wise product is dramatically different for the cells with extreme values of cycle life; note that the cell with a cycle life of 300 cycles (dark red) is almost the mirror image of the cell with 2160 cycles (dark blue), and the trends in these element-wise products consistently match the trends in cycle life (i.e., we consistently observe color gradients from red to blue or vice versa). Furthermore, different voltage regions have different relationships with this element-wise product. Above ~3.35 V, the element-wise product is dominated by noise, as no trend is evident as a function of cycle life. Between ~3.2 V and ~3.35 V, the element-wise product frequently changes sign, but the trends appear meaningful as the magnitude of the element-wise product is consistently ordered by cycle life. The peak at ~3.23 V is sharply negative for the cell with a cycle life of 300 cycles (dark red) and positive for the other example cells; this peak corresponds to the sharp shoulder at ~3.2 V and is consistent with our findings from the univariate models. With the exception of sharply positive regions centered at around 2.4 V and 3.15 V, the element-wise product for the cell with a cycle life of 300 cycles (dark red) is mostly negative below 3.2 V. A consistently signed region appears from around 2.5 V to 3.1 V. This region appears to be among the most sensitive for predicting cycle life; for instance, the cell with a cycle life of 1434 cycles (light blue) closely tracks the cell with a cycle life of 2160 cycles (dark blue) except in this region. At lower voltages, the large positive peak at 2.4 V for the cell with a cycle life of 300 cycles (dark red) corresponds to a small shoulder present in $\Delta Q_{100-10}(V)$ for nearly all cells.

While our models provide some insights into the relationship between elements of $\Delta Q_{100-10}(V)$ and cycle life, we are unable to precisely explain the degradation modes that correspond to each region of the coefficient vector. For instance, on a high level, we are unable to explain why the element-wise product is not consistently negative for cells with low cycle life and consistently positive for cells with high cycle life. These high-rate discharge capacity-voltage curves are inherently challenging to interpret due to practical complications such as peak overlap[38], heterogeneous aging[34], and other effects that convolute the voltage response. However, the consistency of the linear coefficient vectors from the four statistical learning methods, the consistency of the trends with cycle life in these element-wise products, and the relationship between regions of the element-wise product and regions of $\Delta Q_{100-10}(V)$ give us confidence that these coefficient vectors are meaningful. One interesting observation is that the element-wise product for the cell with a cycle life of 300 cycles (dark red) includes (a) a region with a large negative peak centered at ~3.2 V with a magnitude of −0.4; (b) a region with a large negative peak centered at ~3.0 V with a magnitude of −0.2; and (c) a region with a large negative peak centered at ~2.4 V with a magnitude of −0.1. At slightly more positive voltages, these negative peaks have positive counterparts of comparable magnitude. We speculate that these peaks correspond to shifts in the phase transformation plateaus of graphite, which follow a similar trend (e.g., the capacity of each of the three plateaus roughly doubles with increasing voltage)[45]. The voltage response of the graphite electrode is observable in the voltage curves if loss of active graphite material is a dominant degradation mode, as previously established by Severson et al.[3] This hypothesis is further supported by the fact that the voltage response of these LFP/graphite cells is dominated by graphite.[32,33] However, regardless of the validity of this hypothesis, the insights gleaned from this model are both informative and unexpected and thus demonstrate the value of interpretability for data-driven battery lifetime models. We welcome additional explorations of these observations from the broader battery community.

In summary, the use of the elements of the baseline-subtracted capacity matrix as input features for statistical learning methods is a promising approach for creating accurate and interpretable lifetime models. Physics-based degradation modeling could perhaps provide a deeper understanding of the trends present in these coefficient vectors. Additionally, a similar approach coupled with low-rate capacity-voltage curves may be more readily interpretable, as specific degradation modes can often be identified[32]. Furthermore, the use of differential capacity or differential voltage analysis on $\Delta Q(V)$ vectors may further aid interpretability, though perhaps at the cost of accuracy due to the noise introduced by taking the numerical derivative. Finally, this approach may aid in developing generalizable models, particularly when applied to recent synthetic and experimental datasets that span multiple chemistries and cycling conditions.[48,49]

### Multivariate models from capacity matrices

Lastly, we consider four approaches that attempt to capture information from the entire capacity matrix, as opposed to just $\Delta Q_{100-10}(V)$. However, all of these approaches produced models with test errors comparable or larger than the other models presented in this work. We briefly discuss these approaches and their results here for completeness.

First, we considered horizontal slices of the capacity matrices. While vertical slices of the capacity matrices (including metrics such as $\Delta Q_{100-10}(V)$, which is the difference between two vertical slices) capture information over a range of voltages for fixed cycle number, horizontal slices of the capacity matrices (illustrated in Figure 1d) capture information over a range of cycle numbers for fixed voltage. This approach is motivated by the idea that these linear trends may extrapolate and correlate with cycle life. Here, we first identified the voltage that maximizes the absolute value of the slope (i.e., minimizes $\Delta Q_{100-2}(V)$), found to be 3.003 V. We then fit the horizontal slices to a line and use the slope and intercept as input features into the elastic net. We also explored (a) choosing other voltages for the horizontal slice, (b) "voltage averaging", i.e., local averaging across five neighboring voltages, (c) building models with polynomial features from the slope and intercept (e.g., using the square of the slope), and (d) removing outlier points. However, all of these approaches produced models with higher error to other approaches explored in this work (e.g., representative training, primary test, and secondary test errors of 190, 233, and 269 cycles, respectively). We attribute the poor performance of this approach to the nonlinear degradation of these cells (see Figure 1a of Severson et al.[3]), as well as our results throughout this work that demonstrate the value of capturing the dispersion in $\Delta Q(V)$, which is not captured with so few voltage points used.

Second, we considered models that use the entire capacity matrix for feature generation, using the multivariate models presented in Figure 8. The capacity matrix was first reshaped into a vector and then downsampled to avoid training on 100,000 features (we experimented with different downsampling frequencies). However, our attempts were unsuccessful; the models were highly overfit with near-zero training error and large test error. With such a large, collinear feature set, our statistical learning methods were unable to build meaningful models. In some ways, this result demonstrates the value of manual feature engineering for these high-dimensional input datasets when coupled with statistical learning approaches. However, as we discuss soon, the use of the entire capacity matrix may find utility in training end-to-end machine learning and deep learning models for battery lifetime prediction.

Third, we built multivariate models using elastic net regression using the $\log_{10}$-transformed $\Delta Q_{100-10}(V)$ summary features from Figure 3 (excluding the features requiring the use of the absolute value function). Surprisingly, however, the performance of these models is comparable to the univariate models of Figure 3, with RMSEs of 99, 133, and 185 cycles for the training, primary test, and secondary test sets respectively (0%, 7.3%, −2.6% higher errors than the $\log_{10}$(IQR) model). We attribute this outcome to the multicollinearity of the input features; the median correlation coefficient among these features is 0.996, indicating that these features do not capture additional dimensions of that data that are useful for predicting cycle life.

Finally, we trained a convolutional neural network (CNN) model on the capacity matrix. We chose a CNN architecture given the similarities of capacity matrices to images. Here, the input to the model is the baseline-subtracted capacity matrix for each cell, with cycle 10 chosen as the baseline cycle. As in the previous section using the entire capacity matrix, the voltage dimension is downsampled (16 mV sampling frequency). The particular model architecture has two convolutional layers, three fully connected layers, and ReLU activation functions; the model is trained using Adam[50]. Note that the training time for this model is on the order of hours, as opposed to seconds for typical statistical learning models given comparable computational resources. The loss function is the mean squared error plus a regularization penalty on the parameter weights. The weight of the penalty, the number of iterations, and the learning rate are all optimized using the training data. In an effort to use as much data for training as possible, a validation set was not used; instead, hyperparameters were selected by identifying points of diminishing returns in training performance. Drop-out was also considered but did not result in improved performance. The resulting RMSE values were 17, 72, and 204 cycles for the training, primary test, and secondary test sets, respectively—high performance on the primary test set but moderate performance on the secondary test set. The high performance on the primary test set may be attributable to its high similarity to the training set (i.e., the cells in each set came from the same two batches).

We repeated the analysis described above but with the raw capacity measures, i.e. without subtracting the 10$^{th}$ cycle, and observed substantially decreased performance (RMSEs of 47, 129, and 313 cycles for the training, primary test, and secondary test sets, respectively). This result supports our overall hypothesis that feature engineering is a key consideration for strong performance in data-driven battery lifetime prediction models. We also note that the architecture used in the CNN model was developed for image data, which has similarities to the capacity matrix but also some important differences (e.g., lack of translational invariance). Tailored model architectures may play an important role in the development of deep models for battery applications.

## Summary

Table II displays a summary of some models of interest both from this work and from Severson et al.[3] All models substantially outperform a naive benchmarking model (mean of the training set). Our best univariate model, $\log_{10}(\mathrm{IQR}(\Delta Q_{100-10}(V)))$, slightly outperforms the best univariate model from Severson et al.[3], the "variance" model (primary and secondary test errors of 124 and 190 cycles for the IQR model, and primary and secondary test errors of 138 and 196 cycles for the variance model). The model from this work with the lowest test error is the PLSR model, with primary and secondary test errors of 100 and 176 cycles, respectively. However, this model is outperformed by the "discharge" model from Severson et al.[3], which has primary and secondary test errors of 86 and 173 cycles, respectively. Note that the "discharge" model outperforms the "full" model, which had more features available for training, so perhaps the "full" model suffers from more overfitting than the "discharge" model. Lastly, the deep learning models consistently achieved similar results to simpler statistical learning approaches; our work demonstrates that statistical learning methods can achieve competitive performance with deep learning methods while providing additional advantages in terms of training time and interpretability. Neural architectures tailored to this application may improve the performance of deep learning methods relative to statistical learning methods, but improvements should not be assumed and benchmarking using statistical learning methods should be reported.

**Table II.** Summary of RMSE for all regression models in Severson et al.[3] and from selected models in this work. Overall, the "discharge model" from Severson et al.[3] has the lowest test errors. The CNN results are averaged over ten runs. Note that the outlier cell from the primary test set is excluded from all rows; additionally, the training RMSE for the $\log_{10}(\mathrm{var}(\Delta Q_{100-10}(V)))$ is reported as 103 cycles in Severson et al.[3] but 104 cycles in this work.

| Model | Type | RMSE, train (cycles) | RMSE, primary test set (cycles) | RMSE, secondary test set (cycles) |
|---|---|---|---|---|
| Mean of training set (benchmarking model) | Constant | 327 | 399 | 511 |
| $\log_{10}(\mathrm{var}(\Delta Q_{100-10}(V)))$; Severson et al.[3] "variance model" | Univariate | 104 | 138 | 196 |
| Severson et al.[3] "discharge model" | Multivariate | 76 | 86 | 173 |
| Severson et al.[3] "full model" | Multivariate | 51 | 100 | 214 |
| $\log_{10}(\mathrm{IQR}(\Delta Q_{100-10}(V)))$ | Univariate | 99 | 124 | 190 |
| $\log_{10}(\mathrm{percentile}(\Delta Q_{100-10}(V)))$; Percentiles = 31% to 62% | Univariate | 52 | 109 | 261 |
| $\log_{10}(\Delta Q_{100-10}(V\text{=2.913 V}))$ | Univariate | 109 | 118 | 214 |
| Ridge regression on $\Delta Q_{100-10}(V)$ | Multivariate | 85 | 125 | 188 |
| Enet regression on $\Delta Q_{100-10}(V)$ | Multivariate | 92 | 132 | 196 |
| PCR on $\Delta Q_{100-10}(V)$ | Multivariate | 80 | 97 | 193 |
| PLSR on $\Delta Q_{100-10}(V)$ | Multivariate | 59 | 100 | 176 |
| Random forest regression on $\Delta Q_{100-10}(V)$ | Multivariate | 82 | 140 | 202 |
| MLP on $\Delta Q_{100-10}(V)$ | Multivariate | 32 | 140 | 218 |
| CNN on $\Delta Q(V)$ | Multivariate | 17 | 72 | 204 |

To provide additional insight into the residual distribution, we present the cumulative distribution of the absolute residuals for the training, primary test, and secondary test datasets for the six multivariate $\Delta Q_{100-10}(V)$ models, the variance model from Severson et al.[3], and the CNN model (Figure 10). In these plots, high performing models are those with curves close to the upper left corner. The deep learning models (MLP and CNN) both have the best performance on the training set; this trend is expected as, generally, the training error decreases with model complexity. Most models perform comparably on the primary training set, although the CNN model has noticeably lower error. On the secondary test set, the CNN model has the best performance of all models for the majority of the cells. However, this model also has *very* poor performance for two cells (absolute residuals > 700 cycles), which greatly increases the RMSE. The CNN model has the additional challenge of being the most difficult to interpret, making these errors even more challenging to understand. Overall, these results suggest that deep learning approaches may be suitable for applications in which accuracy is paramount and both training time and interpretability are not; in our view, however, simpler statistical learning approaches are generally preferred to deep learning methods, at least for objectives and datasets similar to ours.

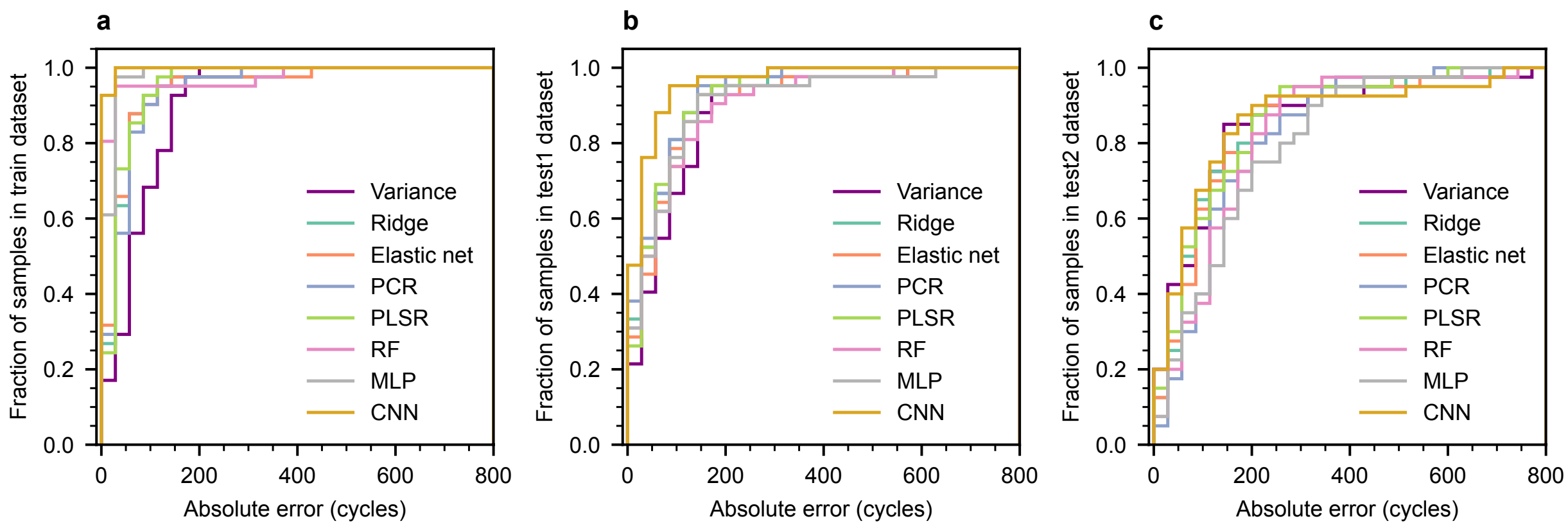

**Figure 10.** Cumulative distribution of the absolute residuals for the (a) training set, (b) primary test set, and (c) secondary test set for the six multivariate $\Delta Q_{100-10}(V)$ models, the variance model from Severson et al.[3], and the CNN model. In these plots, high performing models are those with curves close to the upper left corner. While the performance on the training set generally scales with model complexity, the performance on the test sets is model dependent. Notably, the CNN model outperforms other models on the primary test set, but its moderate performance on the secondary test set is largely caused by two cells with high residuals.

The six features selected in the "discharge" model are the minimum, variance, skewness, and kurtosis of $\Delta Q_{100-10}(V)$, as well as the discharge capacity of cycle 2 and the difference between the maximum discharge capacity and that of cycle 2. One notable feature in this model that is absent from our work is the second-cycle discharge capacity (i.e., the initial capacity, used because the first-cycle capacity was unavailable for one batch). One disadvantage of the baseline-normalized (e.g., baseline-subtracted or baseline-divided) capacity matrix concept is that the original initial capacity values are lost; this simple feature may help capture cell-to-cell differences due to both manufacturing variation and calendar aging. Note that the constant-current capacity of a cycle can be calculated by taking the sum over a column in the unnormalized capacity matrix. Overall, however, the high performance of our simple and interpretable models illustrates the effectiveness of statistical learning approaches for battery lifetime prediction.

## Conclusions

In this work, we designed a general framework for rapid development of data-driven battery lifetime prediction models, using the datasets of Severson et al.[3] and its models as a baseline. Using the voltage curves as our only source of features, we first present the capacity matrix concept for compactly representing and visualizing changes in voltage curves with cycle number. We then explore reducing the voltage sampling frequency, finding that the sampling frequency can be reduced by a factor of 50 without impacting the error. Next, we present a number of univariate models that use summary statistics applied to $\Delta Q_{100-10}(V)$ as input. The univariate interquartile range (IQR) model outperformed the univariate variance model from Severson et al.[3]; we also develop high-performing univariate models using both percentiles of $\Delta Q_{100-10}(V)$ and single elements of the capacity matrix. Additionally, the $\log_{10}$ transformation was consistently found to be an effective transformation for both the features and the cycle life. We then investigate multivariate models that use the elements of $\Delta Q_{100-10}(V)$ as input, with the PLSR method in particular producing a model with low error and interesting insights into the behavior of $\Delta Q_{100-10}(V)$. We also report some approaches that are not effective, including models using multi-cycle averaging and horizontal slices of the capacity matrices. We generally find that the performance of statistical learning methods is comparable more complex deep learning approaches, particularly for generalization, although tailored neural architectures may improve the performance of deep learning models. In summary, the approaches presented in this work produce simple, accurate, and interpretable models for battery lifetime prediction, highlighting the value of domain expertise in feature engineering.

Future work in this space is broad. We hope this work inspires creative feature extraction techniques from capacity matrices. Baseline-normalized capacity matrices can also be applied to other electrochemical data sources like rests and constant-voltage holds. Additionally, as the battery data community develops data storage and representation standards, we propose capacity matrices as one option for compact, machine-learning-ready battery cycling data storage[51]. Statistical learning approaches can also be applied to other objective functions, such as energy-based cycle life, knee point[5], and multipoint prediction[51]. We also recommend applying a similar suite of statistical learning models to benchmark future work that uses new datasets and/or more advanced machine learning methods for battery lifetime prediction, as these approaches provide a reasonable starting point for building high-performing models with minimal human input. Finally, we hope that this work inspires data-driven lifetime prediction models that can generalize to new chemistries and usage conditions; the use of synthetic and experimental datasets that span multiple chemistries and usage conditions[48,49] as training sets may be a step in this direction.

## Acknowledgements

We thank Bruis van Vlijmen, Devi Ganapathi, Marc Deetjen, Norman Jin, Paul Gasper, Sam Greenbank, and Dr. Stephen Harris for insightful discussions.

## Data and code availability

All data and code are publicly available in the associated Zenodo repository[52] and on GitHub (https://www.github.com/petermattia/revisit-severson-et-al).